\documentclass[pdflatex,sn-mathphys-num]{sn-jnl}
\usepackage{graphicx}
\usepackage{float}
\usepackage{booktabs}
\usepackage{multirow}
\usepackage{amsmath,amssymb,amsfonts}
\usepackage{amsthm}
\usepackage[scr=rsfso]{mathalfa} 
\usepackage[title]{appendix}
\usepackage{xcolor}
\usepackage{textcomp}
\usepackage{manyfoot}
\PassOptionsToPackage{hyphens}{url}
\usepackage{listings}
\usepackage{subcaption}         
\usepackage{array,tabularx,makecell} 
\usepackage[final]{microtype}
\usepackage{hyperref}
\theoremstyle{thmstyleone}%

\theoremstyle{thmstyletwo}%

\theoremstyle{thmstylethree}%
\begin{document}

\title[Article Title]{Explainable Knowledge Graph Retrieval-Augmented Generation (KG-RAG) with KG-SMILE}

\author*[1]{\fnm{Zahra} \sur{Zehtabi Sabeti Moghaddam}}\email{zahra.zsm89@gmail.com}
\author[1]{\fnm{Zeinab} \sur{Dehghani}}\email{Z.Dehghani@hull.ac.uk}
\author[1]{\fnm{Maneeha} \sur{Rani}}\email{m.rani3-2022@hull.ac.uk}
\author[1]{\fnm{Koorosh} \sur{Aslansefat}}\email{k.aslansefat@hull.ac.uk}
\author[1]{\fnm{Bhupesh Kumar} \sur{Mishra}}\email{bhupesh.mishra@hull.ac.uk}
\author[1]{\fnm{Rameez Raja} \sur{Kureshi}}\email{r.kureshi@hull.ac.uk}
\author[1]{\fnm{Dhavalkumar} \sur{Thakker}}\email{d.thakker@hull.ac.uk}

\affil[1]{\orgdiv{School of Digital and Physical Sciences}, 
          \orgname{University of Hull}, 
          \orgaddress{\city{Hull}, \postcode{HU6 7RX}, \country{United Kingdom}}}

\abstract{Generative AI, such as Large Language Models (LLMs), has achieved impressive progress but still produces hallucinations and unverifiable claims, limiting reliability in sensitive domains. Retrieval-Augmented Generation (RAG) improves accuracy by grounding outputs in external knowledge, especially in domains like healthcare, where precision is vital. However, RAG remains opaque and essentially a black box, heavily dependent on data quality. We developed a method-
agnostic, perturbation-based framework that provides token and component-level
interoperability for Graph RAG using SMILE and named it as Knowledge-Graph (KG)-SMILE. By applying controlled perturbations, computing similarities, and training weighted linear surrogates, KG-SMILE identifies the graph entities and relations most influential to generated outputs, thereby making RAG more transparent. We evaluate KG-SMILE using comprehensive attribution metrics, including fidelity, faithfulness, consistency, stability, and accuracy. Our findings show that KG-SMILE produces stable, human-aligned explanations, demonstrating its capacity to balance model effectiveness with interpretability and thereby fostering greater transparency and trust in machine learning technologies.}


\maketitle

\section{Introduction}
Recent advances in Generative AI (GenAI), especially Large Language Models (LLMs), have transformed how organisations approach content personalisation, decision-making, and process optimization~\cite{lewis2020retrieval, bechard2024reducing, achiam2023gpt, liang2022holistic}. However, despite their remarkable capabilities, these models still face critical limitations when applied in domains that demand precision and accountability, such as healthcare. One of the most pressing concerns is the tendency of large language models (LLMs) to produce \textit{hallucinations}, outputs that are factually incorrect or unverifiable~\cite{ji2023survey, huang2025survey, lin2021truthfulqa}. Alongside this, ethical risks and embedded biases remain significant barriers to adoption in sensitive settings~\cite{ji2023survey, arrieta2020explainable}.

Retrieval-Augmented Generation (RAG) has emerged as one way to reduce hallucinations. By retrieving external knowledge and combining it with the model's reasoning, RAG helps ground outputs in verifiable facts~\cite{lewis2020retrieval, bechard2024reducing, shuster2022language}. While this improves accuracy compared to standard LLMs, RAG is still essentially a black box; users cannot easily tell which retrieved pieces of information shaped the final response. Moreover, we do not fully understand how these models treat the retrieved knowledge or how they make their final decisions. Knowledge Graphs (KGs) represent entities and their relationships in a structured manner, ensuring consistency and contextual relevance~\cite{zhong2023comprehensive}. When integrated with RAG, this combination, often referred to as GraphRAG, grounds model outputs in a richer and more organised knowledge base. However, despite this structured foundation, GraphRAG still lacks mechanisms to explain how individual graph components contribute to the generated responses.

This study builds on the principles of Explainable Artificial Intelligence (XAI), which emphasise interpretability and accountability in AI outputs~\cite{arrieta2020explainable}. To achieve this, we extend \textbf{SMILE}~\cite{aslansefat2023explaining}, a \textit{model-agnostic}, perturbation-based framework designed to provide fine-grained interpretability for RAG, into \textbf{KG-SMILE}. KG-SMILE works by applying controlled perturbations to the inputs and retrieved knowledge, then measuring the resulting attribution shifts using \textit{Wasserstein distance}, and finally training lightweight \textit{weighted linear surrogate models} to approximate how different KG nodes and edges influence the output. In this way, KG-SMILE identifies the entities and relations most influential to each generated response. This enables GraphRAG not only to produce an answer but also to explain \textit{why} that answer was generated, an essential feature for building trust in high-stakes domains such as healthcare and law~\cite{chalkidis2021lexglue}.

The goal of this study is to improve the transparency and accountability of RAG-based systems. To achieve this, we evaluate the KG-SMILE framework across multiple datasets using attribution-focused metrics, including \textit{fidelity, faithfulness, consistency, stability, and accuracy}~\cite{reimers2019sentence, kusner2015word, colombo2021automatic}. These measures assess the reliability and human-alignment of explanations rather than raw model performance. Our findings show that GraphRAG, equipped with KG-SMILE, produces interpretable, traceable, and trustworthy reasoning traces, effectively balancing the generative strengths of LLMs with the need for accountability.

\subsection{Research Questions}
This research addresses the following questions:
\begin{enumerate}
    \item RQ1: How accurately can a model-agnostic, perturbation-based approach highlight node and edge contributions within a KG for GraphRAG explainability?
    \item RQ2: Which KG components (nodes, edges, relations) have the most significant impact on response stability and attribution accuracy?
\end{enumerate}

The notation and symbols used throughout the paper are summarized in Table~\ref{tab:notation-symbols}.

\begin{table}[ht]
\centering
\renewcommand{\arraystretch}{1.15}
\begin{tabular}{llp{8cm}}
\hline
\textbf{Symbol} & \textbf{Type} & \textbf{Meaning} \\
\hline
$\mathcal{E}$ & Set & Set of entities (KG nodes). \\
$\mathcal{R}$ & Set & Set of relations (KG edge labels). \\
$G$ & Graph & Knowledge graph; $G' \subseteq G$ denotes a subgraph, $G^*$ is the optimal retrieved subgraph. \\
$A$ & Set & Set of candidate answers. \\
$q$ & Query & Natural-language question. \\
$a, a^*$ & Answer & An answer; $a^*$ is the optimal answer to $q$. \\
$T$ & Triple & A KG triple $(e,r,e')$ with $e,e'\in\mathcal{E}$, $r\in\mathcal{R}$. \\
$z$ & Path schema & Relation path $z=\{r_1,\ldots,r_l\}$. \\
$w_z$ & Reasoning path & Instantiation of $z$: $e_0 \xrightarrow{r_1} e_1 \xrightarrow{r_2} \cdots \xrightarrow{r_l} e_l$. \\
$T_q, A_q$ & Sets & Entities mentioned in $q$ and valid answers linked in $G$; $T_q,A_q\subseteq\mathcal{E}$. \\
$P(G)$ & Graph & Perturbed graph produced by removing a set of triples from $G$. \\
$S_{\text{prom}}$ & Index set & Indices of perturbation samples used for evaluation. \\
$\mathbf{R}_{\text{org}}$ & Vector & Response vector (embedding) for the original (unperturbed) output. \\
$\mathbf{R}_{\text{prom}_i}$ & Vector & Response vector for the $i$-th perturbed output. \\
$Emb(\cdot)$ & Function & Text/graph embedding function (produces vectors). \\
$\delta(\cdot)$ & Function & Mapping from text to embedding space (if distinguished from $Emb$). \\
$C_i(\cdot,\cdot)$ & Scalar & Cosine similarity for the $i$-th perturbation. \\
$WD(\cdot,\cdot)$ & Scalar & Wasserstein distance between vectors/distributions. \\
$inv\_WD(\cdot,\cdot)$ & Scalar & Inverse Wasserstein similarity derived from $WD$. \\
$\pi_i$ & Scalar & Similarity-based probability/weight for the $i$-th perturbation. \\
$\beta_0,\ldots,\beta_k$ & Scalars & Regression coefficients in the surrogate linear model. \\
$\epsilon_i$ & Scalar & Error term in the $i$-th regression observation. \\
$\theta,\phi$ & Params & Learnable parameters of retriever and generator. \\
\hline
\end{tabular}
\caption{Symbols used throughout the paper.}
\label{tab:notation-symbols}
\end{table}

\section{Related Works}

This section reviews prior research relevant to our study. We begin with an overview of work in explainable artificial intelligence, highlighting intrinsic and post-hoc approaches. We then examine LIME-based explainability methods, discussing their extensions and adaptations for different domains. Next, we cover efforts in KG and LLM explainability, with a focus on how structured knowledge can improve transparency. We further explore strategies for enhancing explainability through KGs and approaches aimed at addressing hallucinations in LLMs. In addition, we review advances in graph-based interpretability for model decisions and conclude with studies on KG-aided question answering, which directly inform our proposed framework.
\subsection{Explainable Artificial Intelligence}
The rapid advancement in complex machine learning models has heightened the need for explainability to ensure transparency, fairness, and reliability~\cite{adadi2018peeking, doshi2017towards}. Methods can be broadly categorised as intrinsic or post-hoc. Intrinsic methods, such as linear regression or decision trees~\cite{christoph2020interpretable}, are inherently interpretable but often sacrifice predictive power~\cite{rudin2019stop}. Post-hoc methods, such as LIME~\cite{ribeiro2016should} and SHAP~\cite{lundberg2017unified}, explain the decisions of complex "black-box" models without altering their structure~\cite{guidotti2018survey}. Post-hoc approaches can be global, providing insights into overall model behaviour~\cite{du2019techniques}, or local, clarifying individual predictions.

LIME (Local Interpretable Model-Agnostic Explanations) and its derivatives provide case-specific interpretations through simple surrogate models, making outputs accessible even to non-experts~\cite{ribeiro2016should, arrieta2020explainable}. SHAP (SHapley Additive exPlanations), based on Shapley values~\cite{shapley1953value}, offers broader model coverage but is computationally demanding~\cite{lundberg2018consistent, sundararajan2020many}. Other approaches, such as MAPLE~\cite{plumb2018model}, combine interpretable and complex models but introduce additional complexity~\cite{hoffman2018metrics}. Despite their impact, many post-hoc methods still face challenges of stability, fidelity, and scalability. Addressing these limitations has led to the development of advanced frameworks such as SMILE, which aim to provide more reliable and consistent explanations.

\subsection{LIME-based Explainability Methods}
Building on the foundational principles of LIME ~\cite{ribeiro2016should}, numerous methods have been developed to overcome its limitations and extend its applicability across various domains. These approaches target key aspects such as local fidelity, stability, computational efficiency, and domain-specific interpretability~\cite{christoph2020interpretable, guidotti2018survey}. LS-LIME (Local Surrogate LIME) improves fidelity by sampling around regions most relevant to a prediction, albeit with additional computational cost~\cite{laugel2018defining}. BayLIME incorporates Bayesian reasoning and prior knowledge to address kernel sensitivity and inconsistency, thereby enhancing robustness~\cite{zhaobaylime}. SLIME (Stabilised-LIME) applies a hypothesis-testing framework based on the central limit theorem to determine the required number of perturbation points for stable explanations~\cite{zhou2021s}. S-LIME adapts explanations for AI-driven business processes by accounting for feature constraints and task dependencies~\cite{upadhyay2021extending}. OptiLIME balances stability and fidelity through optimised sampling~\cite{visani2020optilime}, while ALIME (Autoencoder-LIME) leverages autoencoders as weighting functions to better approximate decision boundaries~\cite{shankaranarayana2019alime}.

US-LIME improves consistency by generating samples closer to the decision boundary~\cite{saadatfar2024us}. DLIME replaces random perturbations with hierarchical clustering and KNN to ensure determinism~\cite{zafar2021deterministic}. Anchor introduces rule-based "anchors," subsets of features that reliably explain model decisions in critical applications~\cite{garreau2020explaining}. Domain-specific variants have further extended LIME. Sound-LIME~\cite{mishra2017local} focuses on music data, while G-LIME (Global-LIME)~\cite{li2023g} combines global Bayesian regression with ElasticNet-based refinements for deep models. GraphLIME~\cite{huang2022graphlime} explains graph-structured data, though at a high computational cost. Bootstrap-LIME (B-LIME)~\cite{abdullah2023b} adapts LIME for temporally dependent signals such as ECGs through heartbeat segmentation and bootstrapping, thereby improving fidelity and stability.

Recent work has also introduced task-oriented enhancements. DSEG-LIME~\cite{knab2024dseg} supports image segmentation by perturbing object-level regions. SLICE~\cite{bora2024slice} and Stratified LIME~\cite{rashid2024using} refine sampling strategies to improve diversity and subpopulation fidelity. SS-LIME~\cite{lam2025local} integrates self-supervised learning to highlight semantically meaningful regions in the input, offering richer explanations. These advances, along with others~\cite{knab2025lime}, highlight the growing sophistication of post-hoc explainability approaches tailored to deep learning. Finally, SMILE~\cite{aslansefat2023explaining} addresses inconsistency and sensitivity using statistical measures based on Empirical Cumulative Distribution Functions (ECDF). It has been successfully applied across domains such as instruction-based image editing~\cite{dehghani2024mappingmindinstructionbasedimage}, large language models~\cite{dehghani2025explaininglargelanguagemodels}, and point cloud networks~\cite{ahmadi2024explainability}. Although computationally more demanding, SMILE is more resilient to adversarial manipulation, thereby enhancing the trustworthiness of explanations.

Despite these advances, many LIME-based methods still face challenges of scalability, efficiency, and robustness. These limitations motivate our extension of SMILE into KG-SMILE, which leverages structured knowledge graphs to deliver more transparent and reliable explanations.

\subsection{KG and LLM Explainability} 
The combination of KGs and LLMs has become a powerful way to improve the transparency and explainability of AI systems. LLMs, while impressive in their reasoning and generative abilities, often face challenges like factual inaccuracies, hallucinations (where the model generates incorrect or made-up information), and a lack of clarity in how decisions are made~\cite{pan2024unifying}. On the other hand, KGs store structured, fact-based information, which makes them ideal for helping LLMs overcome these limitations. By integrating KGs with LLMs, we can improve how these models retrieve and present information, making them more reliable and easier to understand~\cite{li2024simple}.
This combination leads to more transparent, accurate, and traceable reasoning, addressing some of the core issues that LLMs face on their own~\cite{li2024simple}. This integration helps to improve the overall reliability of LLM-generated responses~\cite{sarmah2024hybridrag}. It also contributes to grounding the responses in structured knowledge, enhancing the system's interpretability~\cite{baek2023knowledge}. Furthermore, the use of KGs can help reduce hallucinations in LLMs, making the outputs more trustworthy~\cite{li2025g}.

\subsubsection{Enhancing Explainability Through KGs}
KGs play a crucial role in making LLMs more explainable by acting as external repositories of structured knowledge. During inference (when the model is making decisions), KGs can be queried to provide clear reasoning paths for the model's responses. Unlike LLMs, which rely on their internal, often opaque memory, KGs offer a more transparent way of retrieving information, which enhances traceability and interpretability~\cite{pan2024unifying}. This means that when an LLM generates an answer, KGs can help back up that answer with specific, factual connections and reasoning~\cite{li2024simple}.
There are several methods that combine KGs with LLMs to improve reliability and transparency. One popular approach is Graph Retrieval-Augmented Generation (GraphRAG), which blends the power of vector-based retrieval with structured graph data~\cite{sarmah2024hybridrag}. For instance, the HybridRAG framework combines KGs with vector retrieval to better ground the generated text in context. This approach not only boosts explainability but also makes AI systems more trustworthy by ensuring that responses are based on structured, factual knowledge~\cite{baek2023knowledge}. With this integration, LLMs can generate responses that are easier to interpret and more reliable, significantly reducing the risk of producing hallucinations or unverified information~\cite{balanos2025kgrag}. This approach also contributes to improving the traceability and accuracy of the reasoning process~\cite{wu2025kg}. Furthermore, it enhances the overall transparency of LLMs by ensuring that generated responses are grounded in solid, factual knowledge~\cite{baghershahi2025nodes}. Finally, the use of GraphRAG techniques improves explainability in a wide range of AI applications~\cite{peng2024graph}.
\subsubsection{Addressing Hallucinations in LLMs}
Hallucinations in LLMs, where models generate factually incorrect or misleading content, remain a persistent challenge, particularly in domains requiring high factual accuracy~\cite{ji2023survey}. KG-Enhanced Inference techniques mitigate this issue by incorporating structured KG data during both pre-training and inference. One such approach, SubgraphRAG, retrieves subgraphs of relevant expertise instead of relying solely on raw text-based retrieval, reducing inconsistencies in generated responses~\cite{li2024simple}.
Another technique, Think-on-Graph (ToG), introduces an LLM–KG co-learning approach, where the LLM acts as a reasoning agent, iteratively traversing a KG using beam search to identify the most promising reasoning paths. This method improves knowledge traceability and enhances the reliability of explanations~\cite{sun2023think}. Furthermore, tightly coupled LLM–KG reasoning enables models to dynamically explore entities and relationships within a KG, ensuring that outputs remain justifiable and interpretable~\cite{sun2023think}.

Recent advancements in addressing hallucinations include a causal reasoning framework, where a study introduced a method called causal-DAG construction and reasoning (CDCR-SFT). This supervised fine-tuning framework trains LLMs to explicitly build variable-level directed acyclic graphs (DAGs) and perform reasoning over them. As a result, this approach improved the model's ability to reason causally and reduced hallucinations by 10\% on the HaluEval benchmark~\cite{li2025mitigating}. Another important development is the Rowen method, which enhances LLMs with a selective retrieval augmentation process explicitly designed to address hallucinated outputs~\cite{ding2024retrieve}.

\subsubsection{Graph-Based Interpretability for Model Decisions} Graph-based explainability methods enhance transparency by explicitly modelling relationships between concepts. GraphLIME, for instance, adapts the LIME framework to graph-structured data, using nonlinear feature selection techniques to improve the interpretability of Graph Neural Networks (GNNs)~\cite{huang2022graphlime}. GraphArena, a benchmarking framework, evaluates LLMs on structured reasoning tasks, revealing the extent to which these models can process and leverage graph-based knowledge~\cite{tang2024grapharena}. LLMs can be used as graph-based explainers in two primary ways: \begin{itemize} \item \textbf{LLMs-as-Enhancers:} LLMs refine textual node attributes in graphs, improving downstream tasks such as node classification and relation prediction~\cite{chen2024exploring}. \item \textbf{LLMs-as-Predictors:} LLMs directly infer graph structures and relationships, reducing dependency on traditional graph models but introducing challenges in accuracy and reliability~\cite{chen2024exploring}. 

\end{itemize}

Additionally, the GraphXAIN method introduces a novel approach to explain Graph Neural Networks (GNNs) by generating narratives that highlight the underlying reasoning paths of the models. This method makes the decision-making process of GNNs more interpretable, providing transparent and human-readable explanations~\cite{cedro2024graphxain}.

\subsubsection{KG-Aided Question Answering}
A key application of KG-enhanced explainability is Knowledge Graph-Augmented Prompting (KAPING). This method embeds KG triples within LLM prompts, significantly improving zero-shot QA performance while reducing hallucinations~\cite{baek2023knowledge}. By dynamically integrating structured knowledge into LLM-generated responses, KAPING ensures factual consistency and improves interpretability. In addition, LLM–KG co-learning frameworks enable LLMs to contribute to automated KG construction, embedding, and completion, allowing for continuous updates and dynamic improvement of kGs over time~\cite{pan2024unifying}. The convergence of KGs and LLMs offers a promising avenue for advancing explainable AI. By integrating RAG with structured KGs, graph-based reasoning paradigms, and interactive knowledge retrieval, AI systems can achieve greater transparency, reliability, and interpretability. However, key challenges remain in scalability, computational efficiency, and generalisability, requiring further research to fully harness the potential of KG–LLM hybrid approaches in real-world applications.

Recent research in this domain includes efforts to mitigate hallucinations in data analytics. One study introduced and evaluated four targeted strategies: Structured Output Generation, Strict Rules Enforcement, System Prompt Enhancements, and Semantic Layer Integration, all aimed at reducing hallucinations in LLMs within data analytics applications~\cite{gumaan2025theoretical}. Another important development comes from Anthropic researchers, who discovered "persona vectors"
, specific patterns of activity within an AI's neural network. These persona vectors influence the AI's behaviour and character traits, potentially explaining why AI responses can unpredictably shift, leading to hallucinations~\cite{chen2025persona}.
The convergence of KGs and LLMs presents an exciting opportunity to advance explainable AI. By integrating retrieval-augmented generation (RAG) with structured KGs, along with graph-based reasoning paradigms and interactive knowledge retrieval, AI systems can offer greater transparency, reliability, and interpretability. However, challenges remain in areas like scalability, computational efficiency, and generalizability, which require further research to unlock the full potential of KG–LLM hybrid approaches in real-world applications.

\section{Problem Definition}
Despite significant advancements in GenAI~\cite{brown2020language, achiam2023gpt} and RAG~\cite{lewis2020retrieval, izacard2022few}, current systems still encounter significant challenges in high-stakes, domain-specific contexts. In fields such as healthcare, precision, traceability, and reliability are essential~\cite{surden2021machine}. However, existing RAG models often struggle to trace the origins of their generated responses, making it difficult to verify their reasoning~\cite{nakano2021webgpt, manakul2023selfcheckgpt}. Moreover, the inability to quantify the contribution of individual data components within external knowledge sources undermines both trust and usability~\cite{bommasani2021opportunities}.

The integration of KGs with RAG, forming what is commonly referred to as GraphRAG~\cite{wang2019deep, edge2024local, barry2025graphrag}, offers an opportunity to enhance contextual relevance and traceability. Nonetheless, substantial gaps remain:

\begin{enumerate}
    \item \textbf{Component-Level Impact:} The influence of specific entities and relationships within a KG on attribution (ATT) fidelity, accuracy, and stability has not been systematically studied~\cite{li2019relation}.
    \item \textbf{Evaluation Metrics:} Reliable approaches for assessing the performance and explainability of GraphRAG systems, particularly under dynamic conditions, are still under development~\cite{doshi2017towards, hoffman2018metrics}.
    \item \textbf{Explainability:} Although GraphRAG systems theoretically improve traceability, few practical methods exist for quantifying and visualising this explainability~\cite{christoph2020interpretable}.
\end{enumerate}

This research, therefore, addresses the following needs:
\begin{itemize}
    \item Developing methodologies to systematically evaluate the impact of individual KG components on AI-generated responses~\cite{peng2024graph}.
    \item Establishing robust metrics, such as cosine similarity and inverse Wasserstein distance, to quantify ATT fidelity and stability~\cite{bhatt2020evaluating}.
    \item Enhancing explainability through perturbation-based analysis and regression modelling, enabling users to interpret better and trust the system's outputs~\cite{ribeiro2016should, plumb2018model}.
\end{itemize}

By addressing these challenges, this study seeks to close critical gaps in the design and evaluation of explainable AI systems, ensuring robustness, transparency, and reliability for domain-specific applications such as legal research~\cite{arrieta2020explainable}.

\section{Proposed Method}
GraphRAG is a framework that leverages external structured KGs to improve the contextual understanding of LLMs and generate more informed responses~\cite{lewis2020retrieval}. The goal of GraphRAG is to retrieve the most relevant knowledge from databases, thereby enhancing the answers of downstream tasks. The process can be defined as Eq.~\eqref{eq:1}~\cite{li2022dynamic}:

\begin{equation}
a^* = \arg \max_{a \in A} p(a|q, G),
\label{eq:1}
\end{equation}

where \( a^* \) is the optimal answer to query \( q \) given graph \( G \), and \( A \) is the set of possible responses. We then jointly model the target distribution \( p(a|q, G) \) with a graph retriever \( p_{\theta}(G'|q, G) \) and an answer generator \( p_{\phi}(a|q, G') \), where \( \theta \) and \( \phi \) are learnable parameters. By applying the law of total probability, \( p(a|q, G) \) can be decomposed as Eq.~\eqref{eq:2}~\cite{sun2019pullnet}:

\begin{equation}
p(a\mid q, G) =
\sum_{G' \subseteq G} p_{\phi}(a\mid q, G')\, p_{\theta}(G'\mid q, G)
\approx p_{\phi}(a\mid q, G^*)\, p_{\theta}(G^*\mid q, G),
\label{eq:2}
\end{equation}

where \( G^* \) is the optimal subgraph. Because the number of candidate subgraphs can grow exponentially with graph size, efficient approximation methods are required~\cite{sudhahar2019reasoning, yasunaga2021qa}. In practice, a graph retriever is employed to extract the optimal subgraph \( G^* \), after which the generator produces the answer based on the retrieved subgraph.

\medskip
The first step of this project entails generating responses to queries using both original and perturbed KGs. The similarity between these responses is then calculated using metrics such as inverse Wasserstein distance~\cite{kusner2015word} and cosine similarity~\cite{mikolov2013efficient} to assess how removed sections affect the system's outputs. In the next step, a simplified regression model is trained on the perturbation vectors, similarity scores, and assigned weights to evaluate the importance of different KG components~\cite{ribeiro2016should}. The coefficients provided by the model indicate how each part of the graph influences the generation process, thereby revealing which sections of the graph are most critical in response generation~\cite{serrano2019attention}. This not only provides valuable insight into the way the model operates but also enhances transparency and trust in GraphRAG's outputs.

Finally, the KG is visualised for more straightforward interpretation. Different nodes and edges are assigned varying colour intensities to indicate their relative importance: the more intense the colour, the greater the contribution of that component. A colour bar is included as a legend to clarify the meaning of these intensities. Such visualisation highlights the internal mechanisms of the system and enhances interpretability by illustrating how each section of the graph impacts the model's responses~\cite{selvaraju2017grad, ying2019gnnexplainer}.

\textbf{Step 1 – Constructing a KG}

To construct the KG for this study, diabetes-related knowledge was extracted from the test partition of the PrimeKGQA dataset, a comprehensive biomedical KGQA resource derived from PrimeKG, the largest precision medicine-oriented KG containing structured triples in subject–predicate–object format~\cite{yan2024primekgqa}. The test dataset, loaded using Python's \texttt{json} library, was filtered to isolate triples relevant to insulin-dependent diabetes mellitus by applying a scoring function that prioritized core terms such as "diabetes," "insulin," and "glucose," alongside secondary terms like "hyperglycemia" and "diabetic neuropathy."  

An undirected graph was constructed using the \texttt{networkx} library, with nodes representing entities (e.g., "Neonatal insulin-dependent diabetes mellitus," "diabetic peripheral angiopathy") and edges denoting predicates (e.g., "associated\_with"). The top 10 connected components were selected to focus on significant diabetes-related subgraphs, which were partitioned into thematic parts (e.g., Part 1 for insulin-related processes, Part 10 for complications), with each part assigned a range of triple indices.  

This structured KG, tailored for integration with a RAG system, enables precise querying and analysis of complex biomedical relationships. The resulting KG serves as a foundational dataset for evaluating the stability and attribution (ATT) fidelity of graph-based representations, which are central to the broader objectives of this research~\cite{paulheim2016knowledge, nickel2015review}.

\medskip
KGs contain abundant factual knowledge in the form of triples, as shown in Eq.~\eqref{eq:3}~\cite{nickel2015review}:

\begin{equation}
    G = \{(e, r, e') \mid e, e' \in \mathcal{E}, r \in \mathcal{R}\},
\label{eq:3}
\end{equation}

where \(\mathcal{E}\) and \(\mathcal{R}\) denote the set of entities and relations, respectively.  
Relation paths, defined as sequences of relations, are given in Eq.~\eqref{eq:4}~\cite{lin2015learning}: 

\begin{equation}
    z = \{r_1, r_2, \ldots, r_l\},
\label{eq:4}
\end{equation}

where \(r_i \in \mathcal{R}\) denotes the \(i\)-th relation in the path, and \(l\) denotes the length of the path.  
Reasoning paths are instances of a relation path \(z\) in KGs (see Eq.~\eqref{eq:5})~\cite{xiong2017deeppath}:

\begin{equation}
    w_z = e_0 \xrightarrow{r_1} e_1 \xrightarrow{r_2} \cdots \xrightarrow{r_l} e_l,
\label{eq:5}
\end{equation}

where \(e_i \in \mathcal{E}\) denotes the \(i\)-th entity and \(r_i \in \mathcal{R}\) the \(i\)-th relation in the relation path \(z\).  

\textbf{Example 1:} Eq.~\eqref{eq:6} gives a relation path:
 
\begin{equation}
    z = \text{Supervises} \xrightarrow{} \text{Works on},
\label{eq:6}
\end{equation}

And a reasoning path instance could be Eq.~\eqref{eq:7}:

\begin{equation}
    w_z = \text{Person 1} \xrightarrow{\text{Supervises}} \text{Person 2} \xrightarrow{\text{Works on}} \text{GraphRAG},
\label{eq:7}
\end{equation}

Which denotes that "Person 1" supervises "Person 2" and "Person 2" is working on "GraphRAG."

\medskip
KGQA is a typical reasoning task based on KGs. Given a natural language question \(q\) and a KG \(G\), the task aims to design a function \(f\) to predict answers \(a \in A_q\) based on knowledge from \(G\), as in Eq.~\eqref{eq:8}~\cite{saxena2020improving}:

\begin{equation}
    a = f(q, G).
\label{eq:8}
\end{equation}

Following previous works~\cite{sun2019pullnet, hu2019retrieve}, we assume the entities \(e_q \in T_q\) mentioned in \(q\) and the answers \(a \in A_q\) are labeled and linked to the corresponding entities in 
\(G\), i.e., \(T_q, A_q \subseteq \mathcal{E}\).

\begin{figure}[htbp]
  \centering
  \includegraphics[width=\linewidth,height=234mm,keepaspectratio]{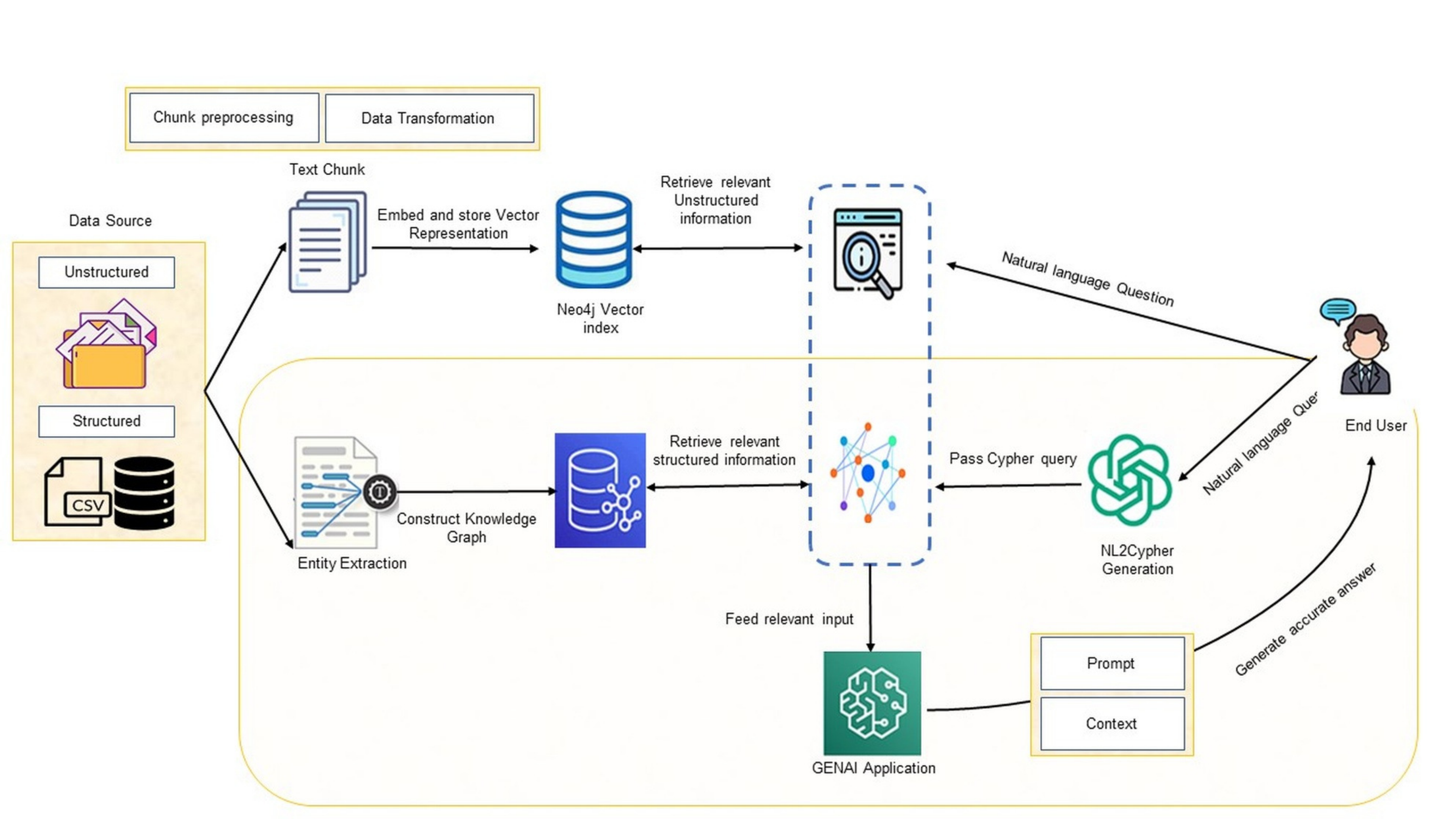}
  \caption{Proposed framework for retrieval of structured and unstructured data with system-formulated queries, showing inputs from structured sources and unstructured documents, processing through a generative AI application, and generation of context-aware and traceable answers.}
  \label{fig:Framework}
\end{figure}

\begin{figure}[htbp]
    \centering
    \includegraphics[width=\linewidth,height=234mm,keepaspectratio]{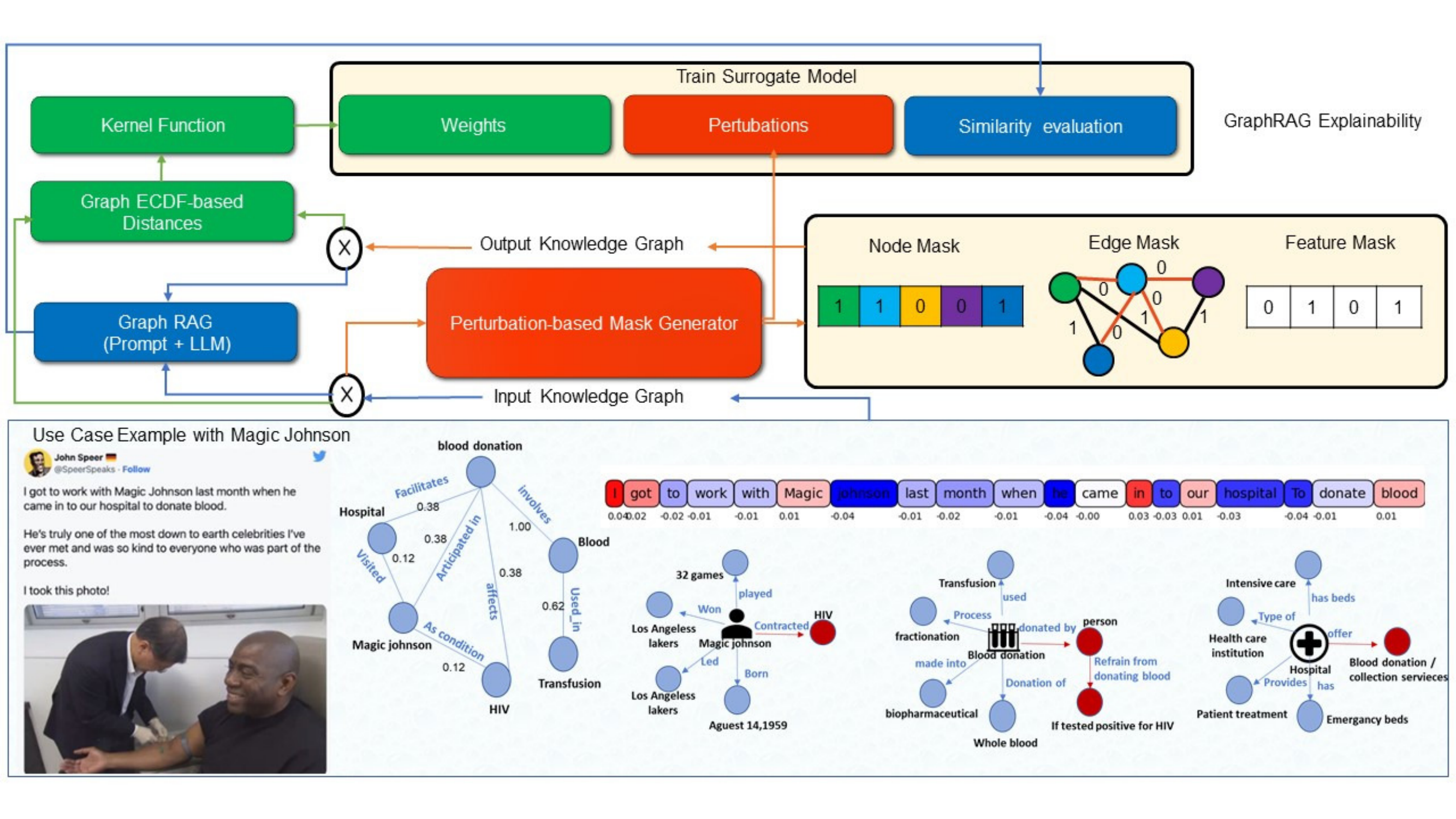}  
    \caption{GraphRAG explainability framework using perturbation-based masks, kernel functions, and ECDF-based distances to train a surrogate model, with node, edge, and feature masks highlighting influential components and illustrating extracted relationships and semantic relevance in healthcare}
    \label{fig:Explainability}
\end{figure}

Alternatively, a predefined knowledge graph can be used. At this stage, accurately defining both nodes and their relationships is essential. Once the knowledge graph is constructed, the prompt is interpreted based on the graph's structure and content. This interpretation generates an answer, which we call the "original answer," serving as a baseline for later improvements, as illustrated in Figs.~\ref{fig:Framework} and \ref{fig:Explainability}.

After receiving a question, we first prompt LLMs to generate several relation paths that are grounded in KGs as plans~\cite{sun2019pullnet, saxena2020improving}. These plans are then used to retrieve reasoning paths from KGs. Finally, faithful reasoning is conducted based on the retrieved reasoning paths, and answers are generated with interpretable explanations~\cite{ying2019gnnexplainer}.

\textbf{Step 2 – Perturbation}

The perturbation process systematically alters the KG to identify influential entities and relationships~\cite{ribeiro2016should, geiger2025causal}. This is achieved by removing specific triplets (entity–relation–entity) individually or in groups, and observing the resulting impact on model performance. Perturbing the graph enables us to clearly pinpoint which components are most critical in generating the original answer.

The perturbation function \( P(G) \) operates by randomly removing sections of the KG \( G \), as defined in Eq.~\ref{eq:9}~\cite{ying2019gnnexplainer}:

\begin{equation}
    P(G) = G - \{ T_{i1}, T_{i2}, \dots, T_{in} \},
\label{eq:9}
\end{equation}

where \( T_{i1}, T_{i2}, \dots, T_{in} \) are the triplets removed during perturbation.  
After perturbation, the system computes similarity between the perturbed and original graphs using cosine similarity between their embedding vectors \( Emb_{org} \) and \( Emb_{prom_i} \), as in Eq.~\ref{eq:10}~\cite{aslansefat2023explaining}:

\begin{equation}
C_i(Emb_{org}, Emb_{prom_i}) = \text{Cosine}(Emb(T_{org}), Emb(T_{prom_i})), \quad \forall i \in S_{prom},
\label{eq:10}
\end{equation}

Where \( Emb(\cdot) \) denotes the embedding function for the respective triplets.

In addition to cosine similarity, Wasserstein Distance (WD) is used to measure the difference between the original and perturbed response vectors, \( \mathbf{R}_{org} \) and \( \mathbf{R}_{prom_i} \). This helps assess how structural perturbations alter the output. The WD is computed as Eq.~\ref{eq:11}~\cite{aslansefat2023explaining}:

\begin{equation}
W_i(\mathbf{R}_{org}, \mathbf{R}_{prom_i}) =
\left( \frac{1}{n} \sum_{j=1}^{n} \left\lVert \delta(Emb(T_{org}^j)) - \delta(Emb(T_{prom_i}^j)) \right\rVert^p \right)^{\!\!1/p}, \quad \forall i \in S_{prom},
\label{eq:11}
\end{equation}

where \( Emb(\cdot) \) is the embedding function and \( \delta(\cdot) \) transforms textual responses into their corresponding embeddings.

By carefully evaluating different metrics, the inverse Wasserstein distance provided the most reliable results for comparing text-based responses generated from pruned graphs~\cite{zhang2025survey, cuturi2013sinkhorn}. A higher similarity score indicates that the removal of specific graph sections has minimal impact on the output. In contrast, a lower score highlights sections that are more influential in shaping the final response.

After experimenting with various perturbation counts (10, 20, 30, 60, and 120), it was determined that 20 perturbations strike the best balance between accuracy and efficiency~\cite{tan2022learning}. Each perturbed graph version \( P(G) \) generates a new response, which is compared to the original response using Wasserstein distance, inverse Wasserstein distance, and cosine similarity for the text output, and cosine similarity for the graph structure. This dual evaluation allows us to measure how perturbations affect both the graph's internal structure and the generated outputs~\cite{ying2019gnnexplainer}.

The calculated distances are used to assign weights to the removed graph components, with a kernel function applied to adjust each component's contribution. These weights highlight the most influential sections of the KG, identifying the parts critical for generating accurate responses~\cite{ribeiro2016should}.

Finally, a probability function \( \pi_i \) based on cosine similarity is applied to analyse the influence of perturbations, as in Eq.~\ref{eq:12}:

\begin{equation}
   \pi_i(Emb_{org}, Emb_{prom}^i) = 
   \exp\left(-\frac{(C_i(Emb_{org}, Emb_{prom}^i))^2}{\sigma^2}\right), 
   \quad \forall i \in S_{prom},
\label{eq:12}
\end{equation}

Where \( C_i \) denotes cosine similarity between the original and perturbed embeddings, and \( \sigma \) is a scaling parameter.  
By combining cosine similarity for the graph structure with inverse Wasserstein distance for text outputs, we provide a comprehensive framework to evaluate the impact of perturbations~\cite{aslansefat2023explaining, ribeiro2016should}.

\textbf{Step 3 – Regression Model for Explainability}

To enhance interpretability, we apply a linear regression model that relates the applied perturbations and their weights to the similarity scores between responses from the original and perturbed KGs~\cite{ribeiro2016should, doshi2017towards}. Let the perturbations be \( P_{i1}, P_{i2}, \dots, P_{ik} \), with corresponding weights \( W_{i1}, W_{i2}, \dots, W_{ik} \). The regression model is defined as Eq.~\ref{eq:13}:

\begin{equation}
   S_i = \beta_0 + \beta_1 W_{i1} P_{i1} + \beta_2 W_{i2} P_{i2} + \dots + \beta_k W_{ik} P_{ik} + \epsilon_i,
\label{eq:13}
\end{equation}

Where:
\begin{itemize}
    \item \( P_{i1}, P_{i2}, \dots, P_{ik} \) are different perturbations applied to the KG,
    \item \( W_{i1}, W_{i2}, \dots, W_{ik} \) are their corresponding weights, reflecting impact on responses,
    \item \( \beta_1, \beta_2, \dots, \beta_k \) are regression coefficients indicating the influence of each weighted perturbation on similarity score \( S_i \),
    \item \( \epsilon_i \) is the error term for the \( i \)-th observation.
\end{itemize}

By mapping weighted perturbations to similarity scores, this regression model quantifies the importance of each KG component. The regression coefficients thus serve as interpretable indicators of how individual nodes and relationships contribute to generating accurate responses, enabling both visualisation and deeper understanding of the model's decision process.

\subsection*{Loss Function}

The regression model minimises the following loss function to determine the best-fit parameters~\cite{hastie2009elements}, as shown in Eq.~\ref{eq:14}:

\begin{equation}
    \text{Loss} = \frac{1}{n} \sum_{i=1}^{n} \left( S_i - \left( \beta_0 + \beta_1 W_i P_i \right) \right)^2,
\label{eq:14} 
\end{equation}

which reduces the squared differences between the actual similarity scores \( S_i \) and the predicted scores from the regression model~\cite{freedman2009statistical}.

\begin{itemize}
    \item \( S_i \) is the similarity score between the original and perturbed graph responses.
    \item \( W_i P_i \) denotes the weighted impact of the perturbation \( P_i \) on the responses.
    \item The regression model links perturbations and their weights to similarity scores, enabling us to determine the relative significance of different graph components in response generation.
\end{itemize}

Based on the regression coefficients, we visualise the most impactful nodes and relationships within the knowledge graph~\cite{christoph2020interpretable}. Varying colour intensities correspond to the relative significance of graph components: the more intense the colour, the greater the contribution to response generation. A colour bar is included to provide a clear legend for interpretation.

This visualisation helps users identify which parts of the knowledge graph contribute most to the system's responses. It also provides transparency by making the system's internal decision-making process more interpretable.

\section{Numerical Results}

\begin{figure}[ht]
    \centering
    \includegraphics[width=\linewidth,height=234mm,keepaspectratio]{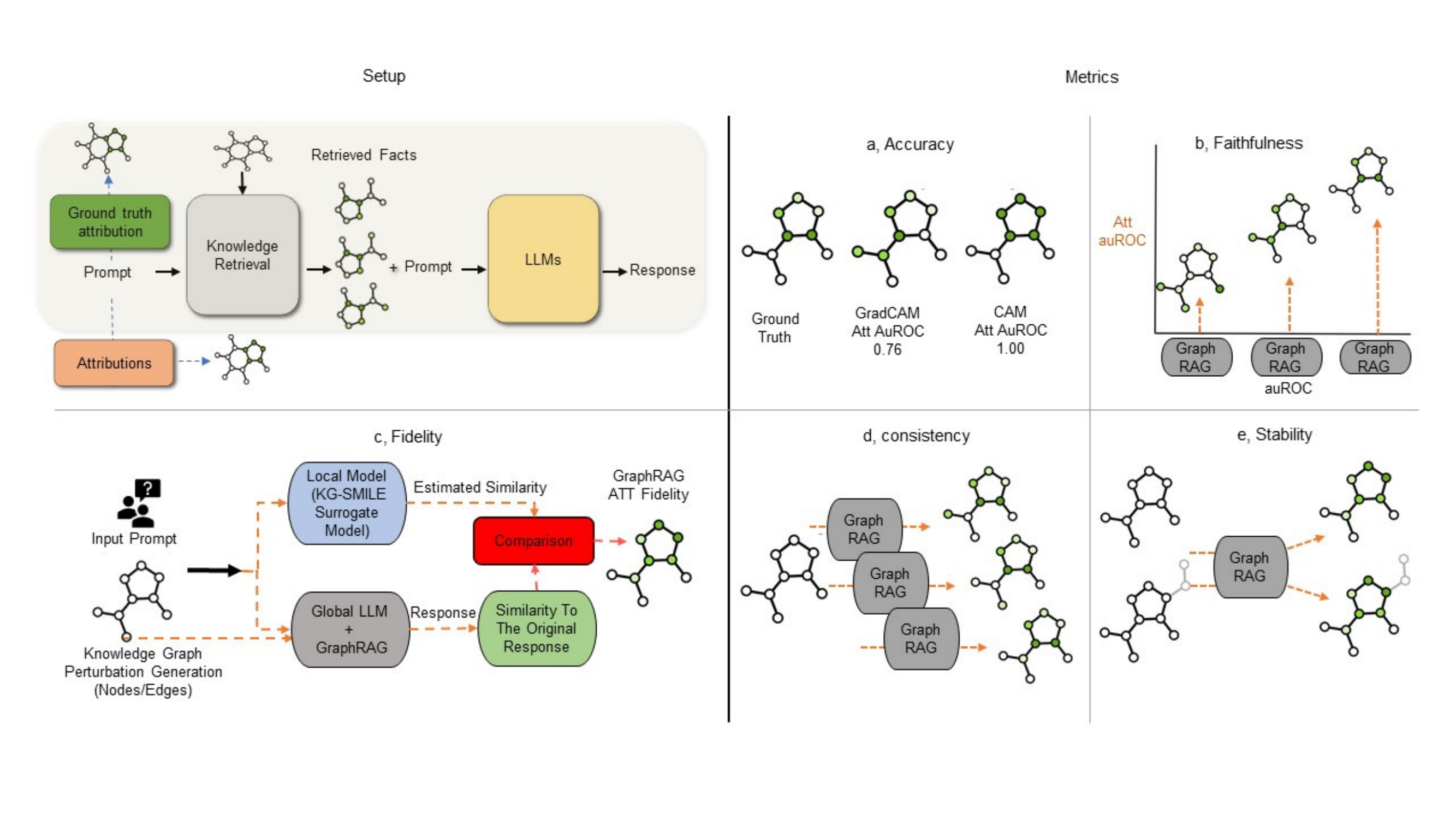}
    \caption{In our investigation, we adopt a suite of evaluation metrics inspired by foundational work provided by Google~\cite{sanchez2020evaluating}, emphasising the multifaceted nature of assessing explainable models. This work highlights the significance of metrics such as accuracy, ATT stability, ATT fidelity and consistency as essential tools for a rigorous evaluation of model behaviour, particularly when comparing explainable models to traditional black-box models. The adoption of these metrics provides a structured methodology to dissect and understand model reliability in a more holistic manner~\cite{sanchez2020evaluating}.}
    \label{fig:Evaluate-Metrics}
\end{figure}

We now present our experimental results regarding the system's attribution (ATT) fidelity, consistency, accuracy, and stability when perturbations are applied to the KG. Weighted linear regression and Bayesian Ridge regression (inspired by BayLIME~\cite{slack2021reliable}) were employed for evaluation, along with similarity measures such as cosine similarity~\cite{huang2013learning}, Wasserstein distance, inverse Wasserstein distance (inv\_WD)~\cite{cuturi2013sinkhorn}, and a hybrid of both. 

Our analysis focuses on the similarity of perturbed KGs compared with the original KG structure and their generated responses. Specifically:
\begin{itemize}
    \item \textbf{Fidelity:} the closeness of perturbed answers to the original answers.
    \item 
    \textbf{Faithfulness:} The degree to which the explanation reflects the true internal workings of the model.
    \item \textbf{Consistency:} the uniformity of model behaviour across multiple perturbations.
    \item \textbf{Accuracy:} whether the generated answer remains correct under perturbations.
    \item \textbf{Stability:} the robustness of outputs when small changes are introduced into the KG.
\end{itemize}
These evaluation criteria, summarised in Fig.~\ref{fig:Evaluate-Metrics}, are essential for understanding system behaviour under varying conditions. In the following sections, we provide detailed results showing how well the perturbed KGs retain both structural integrity and response reliability.

\subsection{ATT fidelity Metrics}

In evaluating ATT fidelity, which measures how well the responses generated from the perturbed KG align with the original KG responses, we used cosine similarity~\cite{huang2013learning}, Wasserstein distance (WD), inverse Wasserstein distance (inv\_WD)~\cite{cuturi2013sinkhorn}, and a combination of them. These metrics were applied using two regression models: weighted linear regression and Bayesian Ridge regression. Our goal was to determine the best metric and model combination to ensure that perturbations to the KG do not compromise the system's output ATT fidelity~\cite{christoph2020interpretable}.
ATT fidelity is quantified by the coefficient of determination \( R^2 \), which evaluates how much of the variance in the original response is explained by the perturbed response~\cite{freedman2009statistical}. The formula for \( R^2 \) is given as Eq.\ref{eq:15}:

\begin{equation}
    R^2 = 1 - \frac{\sum_{i=1}^{N_p} (f(Z_i) - g(Z_i))^2}{\sum_{i=1}^{N_p} (f(Z_i) - \overline{f(Z_i)})^2}
\label{eq:15} 
\end{equation}

Where:
- \( f(Z_i) \) is the original response prediction,
- \( g(Z_i) \) is the perturbed response prediction,
- \( N_p \) is the number of perturbations applied,
- \( \overline{f(Z_i)} \) is the mean of the original predictions.

In addition to \( R^2 \), we used Mean L1 and L2 losses to measure the deviation between the original and perturbed responses~\cite{hastie2009elements, ahmadi2024explainability}. The Mean L1 loss calculates the average absolute difference between the predictions in Eq.\ref{eq:16}:

\begin{equation}
    L_1 = \frac{1}{N_p} \sum_{i=1}^{N_p} |f(Z_i) - g(Z_i)|
\label{eq:16} 
\end{equation}

The Mean L1 loss (\(L_1\)) measures the average absolute deviation between the original responses and the perturbed responses, providing a robust metric against outliers. 

Similarly, the Mean L2 loss (\(L_2\)) evaluates the squared differences between predictions, which penalizes larger deviations more heavily, as shown in Eq.~\ref{eq:17}: 

\begin{equation}
    L_2 = \frac{1}{N_p} \sum_{i=1}^{N_p} (f(Z_i) - g(Z_i))^2
\label{eq:17}
\end{equation}

Weighted versions of these metrics, \( L_1^w \) and \( L_2^w \), incorporate kernel-based weights \(w\) to emphasize the relative importance of specific perturbations. They are defined as: 

\begin{equation}
    L_1^w = \frac{1}{N_p} \sum_{i=1}^{N_p} \left( |f(Z_i) - g(Z_i)| \cdot w \right)
\label{eq:18}
\end{equation}

\begin{equation}
    L_2^w = \frac{1}{N_p} \sum_{i=1}^{N_p} \left( (f(Z_i) - g(Z_i))^2 \cdot w \right)
\label{eq:19}
\end{equation}

The weighted coefficient of determination, \( R_w^2 \), extends the classical \( R^2 \) to include these weights and is defined as Eq.~\ref{eq:20}: 

\begin{equation}
  R_w^2 = 1 - \frac{\sum_{i=1}^{N_p} (f(Z_i) - g(Z_i))^2}{\sum_{i=1}^{N_p} (f(Z_i) - \overline{f_w(Z_i)})^2}  
\label{eq:20}
\end{equation}

Since \( R_w^2 \) can be biased upward in small samples, the weighted adjusted coefficient of determination \( \hat{R}_w^2 \) corrects for both sample size and the number of predictors, as in Eq.~\ref{eq:21}: 

\begin{equation}
    \hat{R}_w^2 = 1 - (1 - R_w^2) \left[ \frac{N_p - 1}{N_p - N_s - 1} \right]
\label{eq:21}
\end{equation}

Finally, to capture the overall discrepancy in the mean values of predicted scores, we compute the mean loss, \( L_m \), as in Eq.~\ref{eq:22}:  

\begin{equation}
    L_m = \left| \frac{\sum_{i=1}^{N_p} f(Z_i)}{N_p} - \frac{\sum_{i=1}^{N_p} g(Z_i)}{N_p} \right|
\label{eq:22}
\end{equation}

This metric quantifies the absolute difference between the average original and perturbed predictions, offering an additional perspective on model stability. The results are summarised in Table~\ref{tab:fidelity-results}.

\begin{table}[ht]
\centering
\setlength{\tabcolsep}{3pt}
\renewcommand{\arraystretch}{1}
\small
\begin{tabularx}{\linewidth}{@{}l
  >{\centering\arraybackslash}X
  >{\centering\arraybackslash}X
  >{\centering\arraybackslash}X
  >{\centering\arraybackslash}X
  >{\centering\arraybackslash}X@{}}
\toprule
\textbf{Metric} & \textbf{Cosine} &
\makecell{\textbf{Inv.}\\\textbf{WD}} &
\makecell{\textbf{Inv. WD}\\\textbf{+ Cosine}} &
\textbf{WD} &
\makecell{\textbf{WD}\\\textbf{+ Cosine}} \\
\midrule
\multicolumn{6}{@{}l}{\textbf{BayLIME}} \\
Mean Loss (LM)   & 1.52E$^{-06}$ & 2.71E$^{-07}$ & 2.30E$^{-07}$ & 2.71E$^{-07}$ & 1.53E$^{-06}$ \\
Mean L1 Loss     & 1.17E$^{-05}$ & 2.10E$^{-06}$ & 1.78E$^{-06}$ & 2.10E$^{-06}$ & 1.18E$^{-05}$ \\
Mean L2 Loss     & 3.79E$^{-10}$ & 1.21E$^{-11}$ & 8.70E$^{-12}$ & 1.21E$^{-11}$ & 3.83E$^{-10}$ \\
Weighted L1 Loss & 2.20E$^{-06}$ & 3.93E$^{-07}$ & 3.34E$^{-07}$ & 3.93E$^{-07}$ & 2.21E$^{-06}$ \\
Weighted L2 Loss & 4.24E$^{-11}$ & 1.35E$^{-12}$ & 9.72E$^{-13}$ & 1.35E$^{-12}$ & 4.27E$^{-11}$ \\
$R^2_\omega$     & 1.00E$^{+00}$ & 1.00E$^{+00}$ & 1.00E$^{+00}$ & 1.00E$^{+00}$ & 0.99999998 \\
$\hat{R}^2_\omega$ & 1.00E$^{+00}$ & 1.00E$^{+00}$ & 1.00E$^{+00}$ & 1.00E$^{+00}$ & 0.99999996 \\
\midrule
\multicolumn{6}{@{}l}{\textbf{Linear Regression}} \\
Mean Loss (LM)   & 2.22E$^{-16}$ & 7.77E$^{-16}$ & 6.66E$^{-16}$ & 5.55E$^{-16}$ & 2.22E$^{-16}$ \\
Mean L1 Loss     & 2.05E$^{-16}$ & 9.49E$^{-16}$ & 9.05E$^{-16}$ & 7.22E$^{-16}$ & 2.16E$^{-16}$ \\
Mean L2 Loss     & 6.10E$^{-32}$ & 1.18E$^{-30}$ & 1.08E$^{-30}$ & 7.57E$^{-31}$ & 6.35E$^{-32}$ \\
Weighted L1 Loss & 2.99E$^{-17}$ & 1.88E$^{-16}$ & 2.26E$^{-16}$ & 1.75E$^{-16}$ & 3.99E$^{-17}$ \\
Weighted L2 Loss & 6.65E$^{-33}$ & 1.57E$^{-31}$ & 2.14E$^{-31}$ & 1.63E$^{-31}$ & 1.02E$^{-32}$ \\
$R^2_\omega$     & 1.00E$^{+00}$ & 1.00E$^{+00}$ & 1.00E$^{+00}$ & 1.00E$^{+00}$ & 1.00E$^{+00}$ \\
$\hat{R}^2_\omega$ & 1.00E$^{+00}$ & 1.00E$^{+00}$ & 1.00E$^{+00}$ & 1.00E$^{+00}$ & 1.00E$^{+00}$ \\
\bottomrule
\end{tabularx}
\caption{Fidelity comparison between perturbed and original KG responses using BayLIME and Linear Regression across similarity metrics.}
\label{tab:fidelity-results}
\end{table}

To thoroughly evaluate the ATT fidelity of responses generated from perturbed KGs, we conducted a comprehensive analysis utilising a variety of metrics. These included cosine similarity~\cite{huang2013learning}, Wasserstein distance (WD), inverse Wasserstein distance (inv\_WD)~\cite{cuturi2013sinkhorn}, and combinations. thereof, all assessed using Weighted Linear Regression and Bayesian Ridge Regression~\cite{hastie2009elements, slack2021reliable}, as shown in Table~\ref{tab:fidelity-results}.

Although some linear models, such as cosine, achieved the absolute lowest weighted $L_2$ losses, we deliberately selected the Linear Regression model using inverse Wasserstein Distance (inv\_WD) because the trade-off in numerical performance is negligible and the benefits for explainability are substantial. As shown in Table~\ref{tab:fidelity-results}, Linear cosine attains a weighted $L_2$ of approximately $6.65 \times 10^{-33}$, while Linear inv\_WD has a weighted $L_2$ of approximately $1.57 \times 10^{-31}$. Both are vanishingly small, and all linear models have weighted $R^2$ values essentially equal to 1, indicating a near-perfect fit. Inv\_WD's interpretive advantages outweigh the marginal gap in raw loss. Unlike cosine, which only measures angular similarity between embeddings, inverse Wasserstein distance captures contrastive differences in the underlying distributions. This makes it more sensitive to meaningful distributional shifts and asymmetries in kG structures. Such sensitivity is critical when the objective is \emph{attribution fidelity}: faithfully identifying which perturbed nodes or relations drive output changes.

KGs are heterogeneous and often exhibit skewed distributions, with some dense clusters and many sparse, rare entities. Metrics based purely on similarity can under-represent rare but influential nodes. In contrast, inv\_WD is derived from the Wasserstein family of distances. We used the inverse Wasserstein Distance in
our weighted linear regression model because the standard
Wasserstein Distance tends to identify nodes with the
opposite influence on response generation. Using inv WD,
we aimed to highlight nodes with a direct impact on output.
and is robust to differences in scale and distributional shape. This makes explanations derived from Linear inv\_WD more robust across both common and rare cases, which is particularly valuable in domains such as healthcare, where long-tail examples often have outsized importance.

Moreover, by combining inv\_WD with a linear model, we retain complete model transparency: coefficients directly indicate the contribution of each perturbation to ATT fidelity changes. This aligns with the central aim of explainable AI research, moving beyond numeric performance to produce explanations that are trustworthy, actionable, and generalizable. Finally, inv\_WD helps avoid overfitting to mere embedding similarity patterns. Similarity measures like cosine perform exceptionally well on in-sample data, but may generalise poorly to novel or rare perturbations. Because inv\_WD explicitly models opposing or divergent distributional changes, it provides richer attribution signals that remain valid as the KG evolves. 

In summary, our choice of Linear inv\_WD reflects a deliberate trade-off: we accept a negligible increase in loss in exchange for markedly improved explanatory power, robustness across heterogeneous data, and theoretical alignment with attribution fidelity objectives. This makes Linear inv\_WD a more appropriate and principled choice for our explainable KG reasoning experiments than simply selecting the numerically top-ranked alternative.
~\cite{freedman2009statistical}.

\subsection{ATT Accuracy}
Across the ten questions, the evaluation results offer insights into the explainability of GraphRAG by examining how effectively the model highlights relevant parts of the KG in response to user queries. The ground truth indicates the critical nodes in the graph that should be highlighted (marked as ``1''), while the model predictions under different temperature settings show which nodes are actually emphasized. The analysis reveals that temperature settings significantly affect the model's ability to retrieve and highlight pertinent information from the KG, influencing both explainability and interpretability~\cite{lin2022teaching}.
On average, the model achieves an AUC of \textbf{0.878} for $T{=}0$ and \textbf{0.744} for $T{=}1$ across all questions (see Appendix~Table~\ref{tab:Auc-table-biomedical} for the detailed per-question results).

\begin{figure}[ht]    
    \centering   
    \includegraphics[width=\linewidth,height=234mm,keepaspectratio]{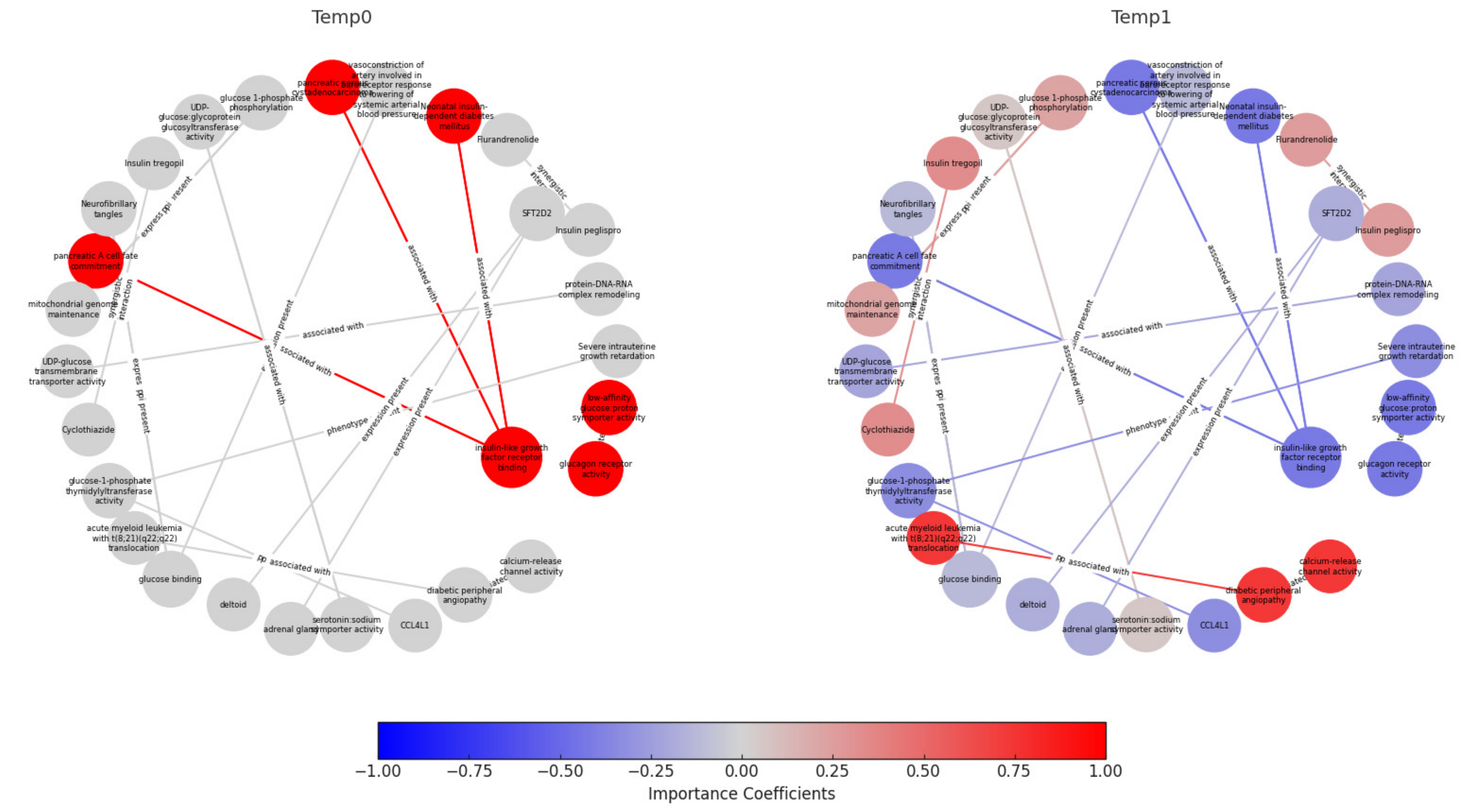}     
    \caption{Accuracy Comparison for the Query ``What is insulin-like growth factor receptor binding associated with?''.}     
    \label{fig:both_figures}
\end{figure}

At a temperature of 0, the retrieval process is deterministic, which means that the model consistently selects the most relevant nodes based on the query. In most instances, the predicted highlights align perfectly with the ground truth (AUC = 1), indicating that the retrieval mechanism successfully identifies and selects the most critical components of the knowledge graph. For example, in queries such as:
\begin{itemize}
    \item \textit{What is insulin-like growth factor receptor binding associated with?}
    \item \textit{What is pancreatic serous cystadenocarcinoma associated with?}
    \item \textit{What is glucagon receptor activity associated with and how it interacts through ppi with low-affinity glucose:proton symporter activity?}
    \item \textit{What is glucose binding?}
\end{itemize}
the model accurately highlights the intended nodes, as shown in Fig.~\ref{fig:both_figures} for temperature 0, demonstrating strong alignment with the ground truth. This level of consistency suggests that deterministic retrieval ensures reliable knowledge attribution, making it especially suitable for tasks where transparency and factual accuracy are essential.

In contrast, at a temperature of 1, the LLM generates responses, introducing variability in how information is processed. Instead of strictly adhering to the retrieved KG components, the LLM may shift its focus to different nodes, sometimes misaligning with the expected ground truth highlights. This misalignment is evident in several cases:

\begin{itemize}
    \item In \textit{What has a synergistic interaction with Flurandrenolide drug?}, the LLM emphasizes incorrect node 'neurofibrillary tangles' alongwith correct node 'insulin peglispro', resulting in an AUC of 0.89.
    \item In \textit{What is UDP-glucose:glycoprotein glucosyltransferase activity associated with and how it interacts with serotonin:sodium symporter activity through ppi?}, again picks the least relevant nodes, leading to an AUC of 0.78.
    \item In \textit{What is Insulin tregopil and with which drug does it synergistic
    interaction?}, the LLM highlights equal correct and incorrect nodes, causing the AUC to drop to 0.56, indicating a complete deviation from the expected explanation.
\end{itemize}

These variations suggest that at higher temperature settings, the model may rely on a broader context rather than strictly following the retrieved knowledge, potentially introducing hallucinations or misattributions~\cite{ji2023survey}. However, not all queries are equally affected. For instance, \textit{What is pancreatic serous cystadenocarcinoma associated with?} and \textit{What is glucagon receptor activity associated with and how it interacts through ppi with low-affinity glucose:proton symporter activity?} maintain perfect alignment (AUC = 1) across both temperature settings, indicating that in these cases, the LLM does not introduce additional noise and remains true to the retrieved knowledge. Conversely, for queries like \textit{What is glucose binding?} and \textit{"What is glucose 1-phosphate phosphorylation and mitochondrial genome maintenance, how do they interact through ppi?}, the variability at temperature 1 results in a drop in AUC to 0.89 and 0.78, respectively, highlighting the inconsistencies introduced by the LLM during explanations.

The observations align with the broader role of temperature settings in LLMs, where the temperature parameter controls the level of randomness in the generated text. This parameter determines how likely the model is to select less probable words instead of the most probable ones~\cite{radford2019language, holtzman2019curious}. The impact of temperature settings on text generation is well-documented:

\begin{itemize}
    \item \textbf{Temperature = 0:} The model consistently picks the most probable next word, resulting in deterministic output. This leads to consistent and factual responses, making it ideal for applications that require high accuracy and predictability. However, responses may sometimes be repetitive or lack creativity.
    \item \textbf{Temperature = 1:} The model introduces more randomness, allowing it to select from a broader range of probable words. This produces more diverse, creative, and human-like responses, which are suitable for applications like storytelling, brainstorming, and creative writing. However, this increased diversity can sometimes compromise coherence and factual accuracy.
\end{itemize}

These effects illustrate how temperature settings influence the explainability of GraphRAG. A lower temperature ensures that the retrieval process is deterministic, leading to consistent and explainable highlighting of relevant KG components. In contrast, generating responses at a higher temperature introduces variability, which can lead to mismatches between retrieved and highlighted knowledge. Although this variability can enhance response diversity in some contexts, it can also compromise explainability, making it more challenging to trace the generated answer back to the kG.

For applications where explainability and factual alignment are critical, such as healthcare scenarios, scientific research, and knowledge-intensive question answering, a lower temperature setting is preferable. This approach ensures that the model reliably highlights the most relevant graph components~\cite{ji2023survey, lin2022teaching}.

\subsection{ATT Faithfulness}

ATT faithfulness refers to the extent to which an explanation accurately reflects a model's reasoning process. A faithful attribution assigns high importance to inputs that genuinely drive the model's output, rather than highlighting spurious correlations or irrelevant features. This property is widely recognised as essential for trustworthy interpretability~\cite{lyu2024towards, agarwal2024faithfulness}.
\sloppy

In our study, we operationalise ATT faithfulness for GraphRAG by measuring the correlation between attribution-based metrics produced by \textbf{KG\textendash SMILE} (here, ATT-AUC at different decoding temperatures: at $T{=}0$ retrieval is deterministic, while at $T{=}1$ the LLM introduces variability in its responses) and the externally reported benchmark accuracies of the same model. Concretely, for OpenAI's GPT-3.5-turbo, for each prompt \(t\), we define
\[
\begin{aligned}
x_t &:= \text{external benchmark accuracy of GPT-3.5-turbo on dataset/prompt } t,\\
y_t &:= \text{ATT-AUC score of KG-SMILE for GPT-3.5-turbo on dataset/prompt } t.
\end{aligned}
\]

We then compute the correlation between these values. Importantly, our results show that whenever the model achieves higher benchmark accuracy, the corresponding ATT-AUC scores are also higher, indicating that more faithful explanations accompany better model performance. As summarised in Table~\ref{tab:att_faithfulness_corr}, across $n{=}10$ prompts we observe a strong positive correlation at $T{=}0$ ($r{=}0.933$) and a weak correlation at $T{=}1$ ($r{=}0.070$).

Formally, we compute the Pearson correlation coefficient:
\begin{equation}
r = \frac{\sum_{i=1}^{n} \bigl(x_i - \bar{x}\bigr)\bigl(y_i - \bar{y}\bigr)}{\sqrt{\sum_{i=1}^{n} (x_i - \bar{x})^2}\,\sqrt{\sum_{i=1}^{n} (y_i - \bar{y})^2}},
\label{eq:pearson_kg_bench}
\end{equation}
where $x_i$ are the externally reported benchmark accuracies, $y_i$ are the corresponding KG-SMILE attribution scores (ATT-AUC at $T{=}0$ or $T{=}1$), and $\bar{x}, \bar{y}$ denote their means. A strong positive $r$ indicates that attribution metrics faithfully track external measures of model competence.

\begin{table}[ht] 
\centering
\begin{tabular}{ccc}
\hline
Temperature $T$ & $n$ & Pearson $r$ \\
\hline
0 & 10 & 0.975524 \\
1 & 10 & 0.845926 \\
\hline
\end{tabular}
\caption{Pearson \(\!r\!\) between ATT--AUC \((y_{i})\) and external accuracy \((x_{i})\).}
\label{tab:att_faithfulness_corr}
\end{table}

Thus, in our framework, ATT faithfulness is an externally validated property: explanations are considered reliable to the extent that KG-SMILE ATT-AUC scores correlate with established benchmarks, ensuring that interpretability provides meaningful insights into knowledge-graph–augmented language models, particularly in high-stakes applications~\cite{jacovi2020towards}.

\subsection{ATT Stability}

ATT stability evaluates how well a model maintains its accuracy and consistency in the face of minor perturbations to the KG~\cite{carvalho2019machine}. This property is a vital component of an explainable model, as it ensures that small changes to the input do not significantly affect the model's predictions or the explanations it generates~\cite{ribeiro2016should, slack2021reliable}. 

In this study, ATT stability was assessed by perturbing the KG through slight modifications to its structure, such as altering relationships between nodes, and then measuring the similarity between the original and perturbed explanations. The Jaccard index~\cite{ahmadi2024explainability, burger2025towards}, defined in Eq.~\ref{eq:23}, was used for this purpose:

\begin{equation}
    Jaccard(A, B) = \frac{|A \cap B|}{|A \cup B|}
\label{eq:23}
\end{equation}

Where \( A \) is the set of explanations from the original graph and \( B \) is the set of explanations from the perturbed graph. The index ranges from 0 to 1 and quantifies the similarity between the two sets. Higher values indicate greater ATT stability.

\begin{table*}[h!]
\centering
\renewcommand{\arraystretch}{1.2}
\begin{tabularx}{\textwidth}{>{\raggedright\arraybackslash}X cc}
\toprule
\textbf{Perturbation triple added} & \textbf{Jaccard ($T{=}1$)} & \textbf{Jaccard ($T{=}0$)}\\
\midrule
\small (``insulin-like growth factor receptor binding'', ``associated\_with'', ``\nolinkurl{type_1_diabetes}'') & 1.00 & 0.10 \\
\small (``pancreatic serous cystadenocarcinoma'', ``treated\_with'', ``chemotherapy'') & 0.90 & 1.00 \\
\small (``glucagon receptor activity'', ``inhibited\_by'', ``insulin'') & 0.80 & 1.00 \\
\small (``glucose binding'', ``expression\_present'', ``pancreas'') & 0.70 & 1.00 \\
\small (``Flurandrenolide'', ``used\_for'', ``diabetic\_skin\_conditions'') & 0.80 & 1.00 \\
\small (``glucose 1-phosphate phosphorylation'', ``associated\_with'', ``glycogen\_storage\_disease'') & 1.00 & 1.00 \\
\small (``UDP-glucose:glycoprotein glucosyltransferase activity'', ``regulates'', ``protein\_folding'') & 0.90 & 1.00 \\
\small (``SFT2D2'', ``expression\_present'', ``pancreatic\_islets'') & 1.00 & 1.00 \\
\small (``diabetic peripheral angiopathy'', ``caused\_by'', ``chronic\_hyperglycemia'') & 1.00 & 1.00 \\
\small (``Insulin tregopil'', ``used\_for'', ``type\_2\_diabetes'') & 1.00 & 1.00 \\
\bottomrule
\end{tabularx}
\caption{Perturbation triples and corresponding Jaccard similarities for $T{=}1$ and $T{=}0$.}
\label{tab:s-table}
\end{table*}

The results in Table~\ref{tab:s-table}. indicate that the ATT stability of explanations generated by the knowledge graph (KG) model is significantly influenced by both structural changes and temperature settings~\cite{slack2021reliable, carvalho2019machine}.

\begin{figure*}[ht]
    \centering
    \includegraphics[width=\linewidth,height=234mm,keepaspectratio]{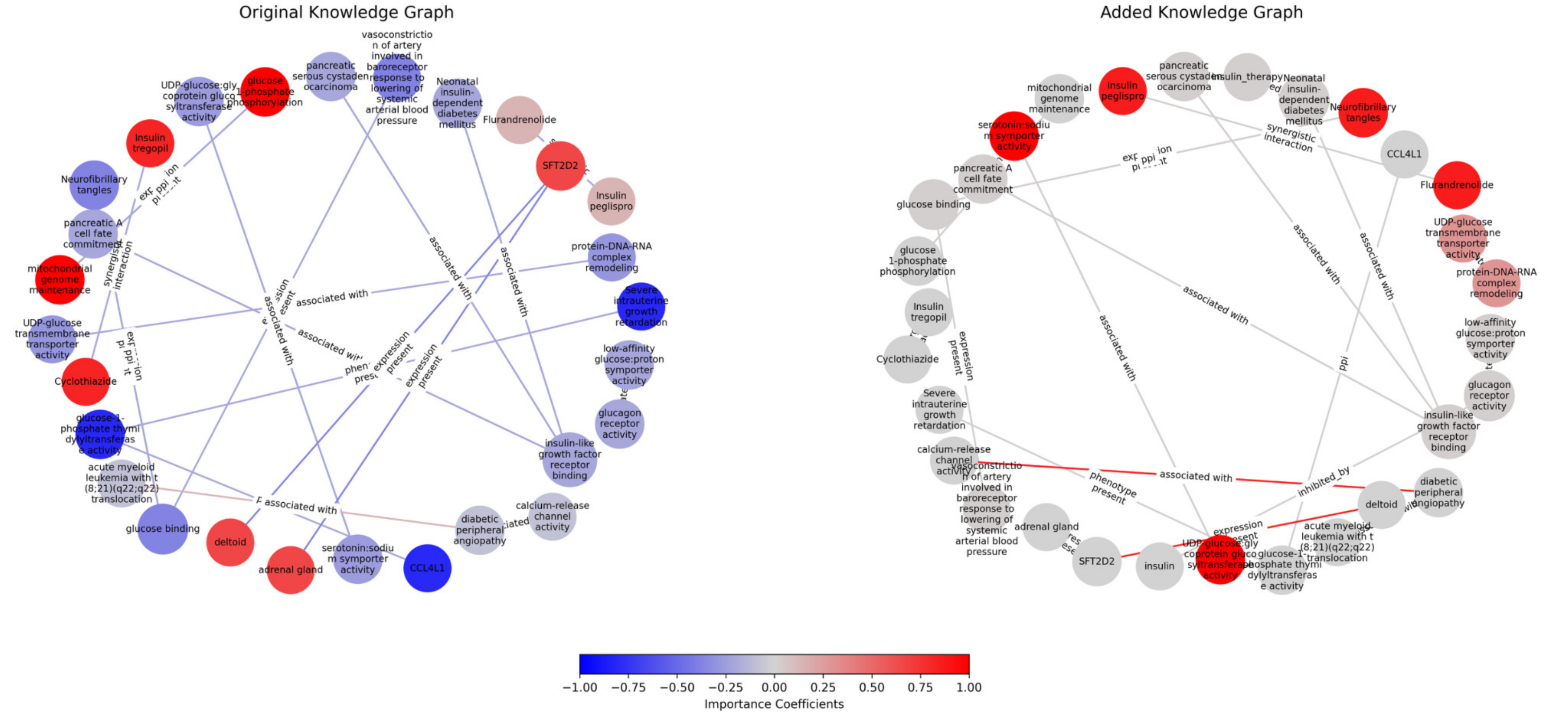}
    \caption{Impact of Adding the Triplet ("glucagon receptor activity", "inhibited by", "insulin") at Temperature 0: Jaccard Similarity 1.0
 in Knowledge Graphs}
    \label{fig:Stability2}
\end{figure*}

As shown in Fig.~\ref{fig:Stability2},
Analysis of temperature settings shows that at $T=0$, where the model operates deterministically, explanations remain stable despite minor modifications to the KG. Here, T denotes the temperature parameter controlling the randomness of the model's output. This is evidenced by consistently high Jaccard similarity values (1.000 in multiple cases), indicating that the model retrieves explanations that are nearly identical to those generated from the original KG. This suggests that, in a low-temperature setting, the model prioritizes fixed reasoning paths, maintaining robustness against small perturbations and ensuring consistency and reliability in its explanations~\cite{ribeiro2016should}.

\begin{figure*}[ht]
    \centering
    \includegraphics[width=\linewidth,height=234mm,keepaspectratio]{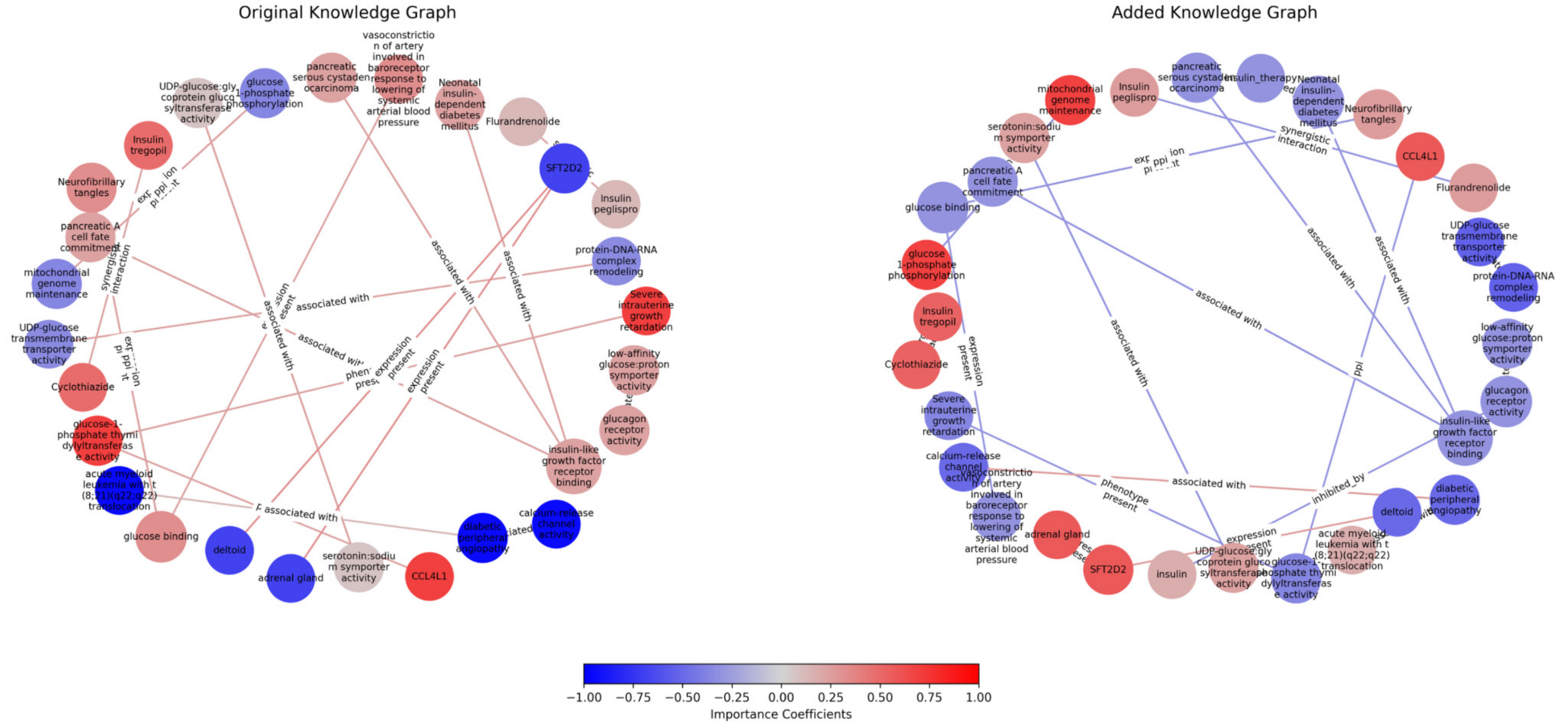}
    \caption{Impact of Adding the Triplet ("glucagon receptor activity", "inhibited by", "insulin") at Temperature 1: Jaccard Similarity 0.8 in Knowledge Graphs}
    \label{fig:stability}
\end{figure*}

Conversely, at $T=1$, where the model introduces stochasticity in its response generation, a slight decline in ATT stability is observed, with Jaccard similarity values dropping a certain values in most of the cases. This indicates that higher temperature settings lead to greater sensitivity to perturbations, prompting the model to explore alternative reasoning pathways and retrieve explanations that deviate significantly from the original~\cite{jiang2019learning}. The introduction of additional triples, even if unrelated to the queried entity, alters the structure of the graph, causing a cascading effect that disrupts the model's ability to maintain consistency in its explanations~\cite{carvalho2019machine}, as illustrated in Fig.~\ref{fig:stability}, which depicts the ATT stability analysis.

This contrast between deterministic behavior at $T=0$ and stochastic behavior at $T=1$ highlights the dual nature of temperature settings in LLM-driven KG reasoning~\cite{lin2022teaching}. While a low-temperature setting ensures ATT stability, a higher-temperature setting enhances adaptability, allowing the model to explore diverse responses but at the cost of increased variability in explanations. The sensitivity observed at $T=1$ suggests that even minor structural modifications in the KG can significantly alter the retrieved explanations, reducing the model's reliability in dynamic environments where perturbations are common. This raises concerns about the trade-off between explainability and adaptability, as a more stochastic approach may enhance response diversity but weaken robustness in explanation retrieval~\cite{slack2021reliable}.

The findings emphasize the importance of temperature control in KG-based reasoning models, as temperature directly affects the balance between stability and exploration. A key implication is that models designed for high-explainability tasks should favor lower temperatures, where reasoning pathways remain consistent even under minor modifications to the KG~\cite{ribeiro2016should, carvalho2019machine}.

\subsection{Consistency}

Consistency assesses how predictable the system is in its responses to small perturbations in the KG. It evaluates how stable and reliable the model's responses are when minor modifications are made to the KG structure ~\cite{jiang2019learning, ribeiro2016should}. In our analysis, we applied slight perturbations across different sections of the KG and evaluated the consistency of the model's responses. Across 50 runs, the responses varied minimally across different partitions of the KG (parts 1 through 10). Part 1 exhibited the least variability, and the remaining parts also showed only minor variations, indicating that the system's responses remained consistent in the face of small changes across all sections ~\cite{slack2021reliable}. The narrow range of variations suggests that the model's responses were stable and largely unaffected by minor perturbations. Despite the slight differences observed, the model demonstrated strong overall consistency by maintaining reliable responses across all parts of the KG. This suggests that the system is highly predictable when minor perturbations are introduced and can handle them without a significant decline in performance. However, if more extensive or systematic alterations are applied to critical nodes in the KG, greater variation could occur. Small, localized changes, however, did not alter the system's high degree of consistency~\cite{carvalho2019machine}.

The results from the 50 runs confirm strong consistency across most sections of the KG and demonstrate that the model can produce stable and predictable responses even when the graph is exposed to minor perturbations.

\subsection{Computation Complexity}

\begin{table*}[ht]
\centering
\renewcommand{\arraystretch}{1.2}
\begin{tabularx}{\textwidth}{>{\raggedright\arraybackslash}X rr}
\toprule
\textbf{Method} & \textbf{BayLIME} & \textbf{Linear Regression} \\
\midrule
Cosine similarity & 25.01s & 25.60s \\
Inverse Wasserstein distance & 20.02s & 25.20s \\
Inverse Wasserstein + Cosine (hybrid) & 23.32s & 23.43s \\
Wasserstein distance & 27.02s & 25.68s \\
Wasserstein + Cosine (hybrid) & 18.18s & 17.17s \\
\bottomrule
\end{tabularx}
\caption{Wall time comparison of ATT fidelity configurations. The graph metric is fixed to cosine similarity; row labels denote the text-side metric.}
\label{tab:walltime_ATTfidelity}
\end{table*}
To systematically evaluate ATT fidelity across text and graph representations, we analyzed multiple similarity metrics, including \textbf{Cosine}~\cite{huang2013learning}, \textbf{Wasserstein + Cosine (hybrid) }~\cite{cuturi2013sinkhorn, peyre2019computational}, \textbf{Inverse Wasserstein distance }~\cite{zhang2025survey}, and \textbf{Inverse Wasserstein + Cosine (hybrid) }~\cite{tan2022learning}. Our experiments, which included \textbf{20 perturbations per configuration}~\cite{ying2019gnnexplainer}, compared the performance of \textbf{BayLIME}~\cite{zhao2021baylime} and \textbf{Linear Regression} models~\cite{freedman2009statistical}, focusing on both \textbf{computational efficiency and ATT fidelity}~\cite{christoph2020interpretable}.

Table~\ref{tab:walltime_ATTfidelity} illustrates the trade-off between computational efficiency and accuracy when analyzing text and graph representations~\cite{ribeiro2016should}. The fastest method, which combines Hybrid Text Metrics (Wasserstein Distance + Cosine) with Graph Metrics (Cosine) ATT fidelity Analysis, achieves the lowest processing time of \textbf{17.17 seconds}. This makes it the most efficient option for rapid evaluations while still maintaining reasonable accuracy, as confirmed by linear regression analysis. In contrast, the BayLIME ATT fidelity Analysis of Text-to-Text using Inverse Wasserstein Distance and Graph-to-Graph using Cosine, takes \textbf{27.02 seconds}. This method entails a significantly higher computational cost due to its use of Inverse Wasserstein Distance, which improves explainability by effectively capturing opposing nodes~\cite{aslansefat2023explaining}.

These results highlight the \textbf{trade-off between computational efficiency and ATT fidelity}, emphasizing the importance of selecting appropriate \textbf{similarity metrics} based on the specific requirements of the task~\cite{lin2022teaching}. Experiments were performed on Google Colab's standard runtime environment, which provides approximately 12GB of RAM, demonstrating efficient time consumption compared to alternative approaches.

\subsection{Chain of Thought}
In complex QA tasks, particularly those involving GraphRAG, traditional LLMs often struggle to produce meaningful responses due to their reliance on direct retrieval and pattern matching~\cite{yao2023tree, zhou2022least}. Many real-world queries require multi-step reasoning, where multiple entities and relationships within a KG must be linked together before arriving at a well-supported conclusion. To address this challenge, we employ Chain-of-Thought (CoT) reasoning~\cite{wei2022chain}, a structured approach that guides the GraphRAG system through intermediate reasoning steps, ensuring that answers are generated in a logical, explainable manner while leveraging the structured representation of the KG.

{Example Question Requiring Chain-of-Thought Reasoning in GraphRAG}
Consider the following complex question:
\begin{quote}
\textit{'Which disease is associated with glucose binding through its link to neurofibrillary tangles, and how does this connect further to a condition related to calcium-release channel activity?'}
\end{quote}
This query requires GraphRAG to conduct multi-step reasoning, as it involves:
\begin{itemize}
    \item Understanding glucose binding glucose binding through its link to neurofibrillary tangles.
    \item Understanding diseases associated with diabetic binding.
    \item Analysing its connection to a condition related to calcium-release channel activity.
\end{itemize}
A simple LLM query often fails to adequately address such questions, as it lacks the ability to connect these concepts explicitly within a structured KG. When simply run, it may return: \textit{``I'm sorry, I don't have enough information to provide a helpful answer.''} Instead, a CoT-based GraphRAG reasoning framework extracts relevant knowledge from the KG, traces logical connections, and formulates an answer grounded in explicit relationships rather than implicit language model heuristics~\cite{wang2022self, zelikman2022star}.

\subsubsection{Extracting and Structuring Relevant KG Triples}
The first step in this approach is extracting key entities and relations from the input question using NLP. The function \texttt{extract\_entities\_relations(question)} identifies relevant entities e.g. 'glucose', 'disease', 'diabetic', 'neurofibrillary tangles' ) and relationships (e.g., bind', 'relate', 'associate', 'interact'). The function \texttt{generate\_chain\_of\_thought(kg, question)} then iterates through subject-predicate-object triples stored in the KG, constructing a reasoning chain such as:

\begin{align*}
    \text{glucose binding} &\rightarrow \text{[expression present]} \rightarrow \text{Neurofibrillary tangles} \\
    \text{Neurofibrillary tangles} &\rightarrow \text{[associated with]} \rightarrow \text{Disease} \\
    \text{Disease} &\rightarrow \text{[associated with]} \rightarrow \text{calcium-release channel activity}
\end{align*}

By extracting and organizing these structured triples, the GraphRAG model enhances reasoning interpretability, ensuring that each step in the answer is traceable within the KG.

Complex questions such as the one above cannot be effectively answered by a direct LLM query alone. Unlike simple retrieval-based approaches, which rely on matching isolated facts, GraphRAG with CoT reasoning ensures a structured approach to logical inference. Without CoT, the system may fail to correctly connect multi-hop relationships, leading to fragmented or incomplete answers. The CoT framework decomposes queries into sequential reasoning steps, ensuring both accurate and explainable responses~\cite{press2022measuring}.

Once the reasoning chain is established, the extracted triples are formatted into a structured prompt using the function \texttt{format\_triples\_for\_prompt(}\allowbreak\texttt{reasoning\_chain)}. This process guides the LLM with explicit, factually grounded information, reduces hallucinations, and ensures responses remain faithful to the KG. By explicitly presenting the logical sequence of facts, the CoT approach enables GraphRAG to generate structured and context-aware answers.

\subsubsection {Explainability Through Perturbation Analysis}

\begin{figure*}[ht]
    \centering
    \includegraphics[width=\linewidth,height=234mm,keepaspectratio]{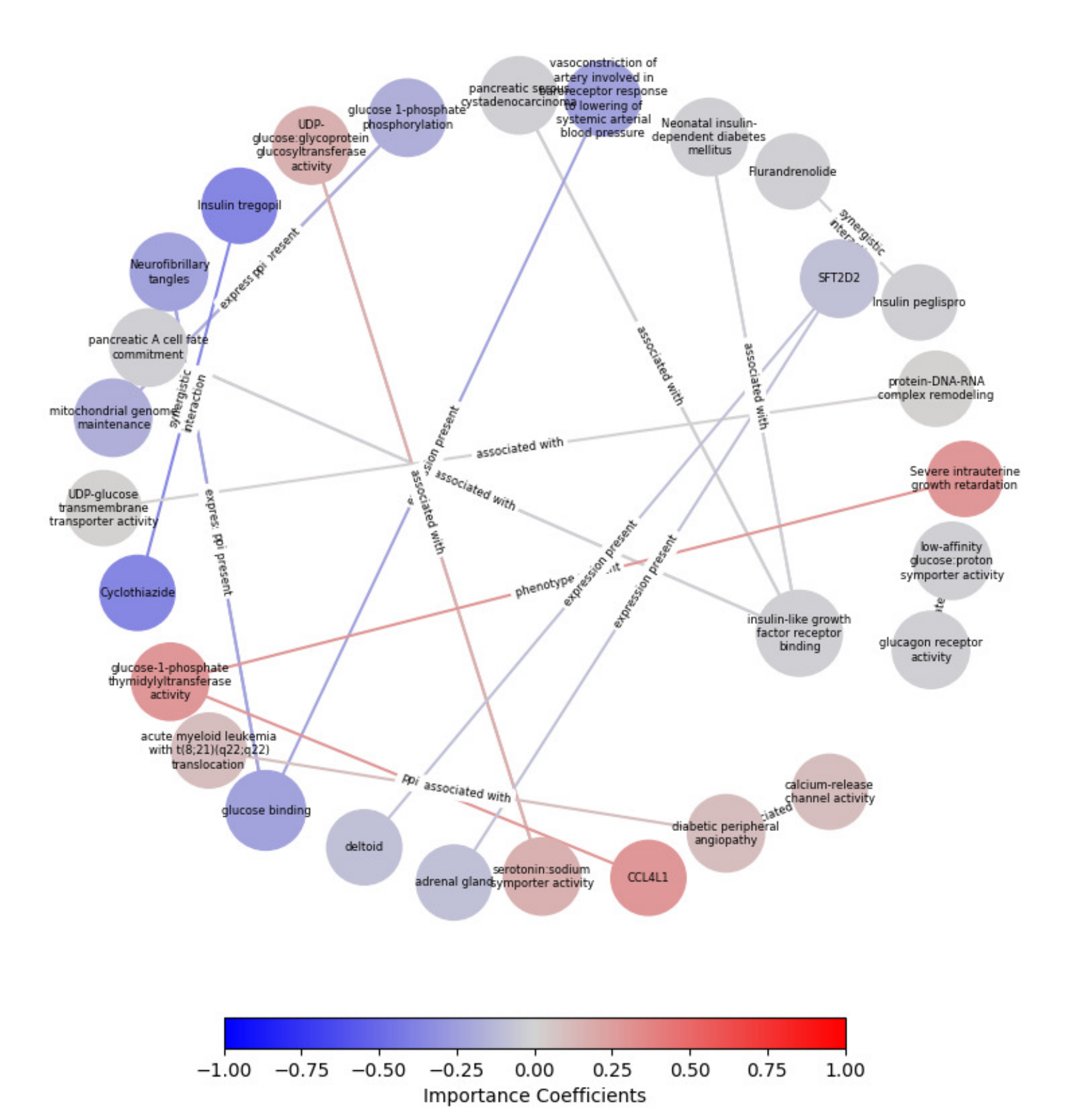}
    \caption{ Chain of Thought KG-SMILE Explainability  }
    \label{fig:COT}
\end{figure*}

To assess the explainability of GraphRAG, we implemented KG-SMILE, a perturbation-based approach that systematically altered the KG structure and evaluated the impact on generated responses. In each iteration, we generated a perturbed version of the KG by selectively removing or modifying nodes and edges. This perturbed KG was then fed into the Chain-of-Thought (CoT) reasoning framework to generate an answer. By comparing the perturbed responses with the original, we measured their alignment using both cosine similarity~\cite{huang2013learning} and inverse Wasserstein distance~\cite{cuturi2013sinkhorn}. This iterative process effectively identified the most influential nodes and relationships~\cite{ying2019gnnexplainer, ribeiro2016should}, revealing their critical role in shaping the final generated response. As illustrated in Fig.~\ref{fig:COT}, the importance coefficients highlight key nodes which significantly contribute to the reasoning. KG-SMILE provides a structured mechanism to interpret the decision-making process of GraphRAG, ensuring greater transparency and explainability in AI-driven question answering.

\subsection{Pre-Prompt Mechanism}

A common challenge in \textbf{KGQA} systems is their occasional inability to retrieve relevant information when queried with a specific phrasing, often resulting in responses such as \textit{"I do not know."} This limitation arises due to the sensitivity of retrieval models to linguistic variations and the inherent constraints of structured knowledge representations. To address this issue, we employ a \textbf{pre-prompting mechanism} that generates multiple \textbf{rephrased versions} of the original query~\cite{dong2017learning, anantha2020open}, increasing the likelihood of alignment with stored knowledge. By broadening the phrasing spectrum, this approach enhances retrieval efficacy and mitigates the risk of response failure.

Once the rephrased questions are generated, they are individually processed through GraphQAChain to obtain possible answers. However, responses that still return \textit{"I do not know"} or similar uncertainty indicators are systematically discarded~\cite{rajpurkar2018know}. A \textbf{filtering mechanism} ensures that only valid, meaningful responses are retained for further analysis. This selective approach significantly enhances response reliability by eliminating uncertainty and ensuring that the final answer is constructed from informative, contextually relevant outputs.

To determine the most accurate response, the filtered answers are aggregated using \textbf{embedding-based similarity techniques~\cite{reimers2019sentence}}, which synthesize the most semantically representative response. By leveraging multiple valid responses rather than a single direct retrieval, this process improves answer robustness and completeness. However, while this method enhances response accuracy, it introduces a trade-off in explainability~\cite{jacovi2020towards}. Because the final answer is an aggregation of multiple responses rather than directly extracted from a single node, the system highlights more nodes in the KG, similar to the way a simple LLM-based retrieval process does but with slightly less intensity. While the correct knowledge component is still identified, the aggregation process results in a broader distribution of emphasis across multiple nodes rather than pinpointing a single definitive source.
\begin{figure*}[tb]
    \centering
    \includegraphics[width=\linewidth,height=234mm,keepaspectratio]{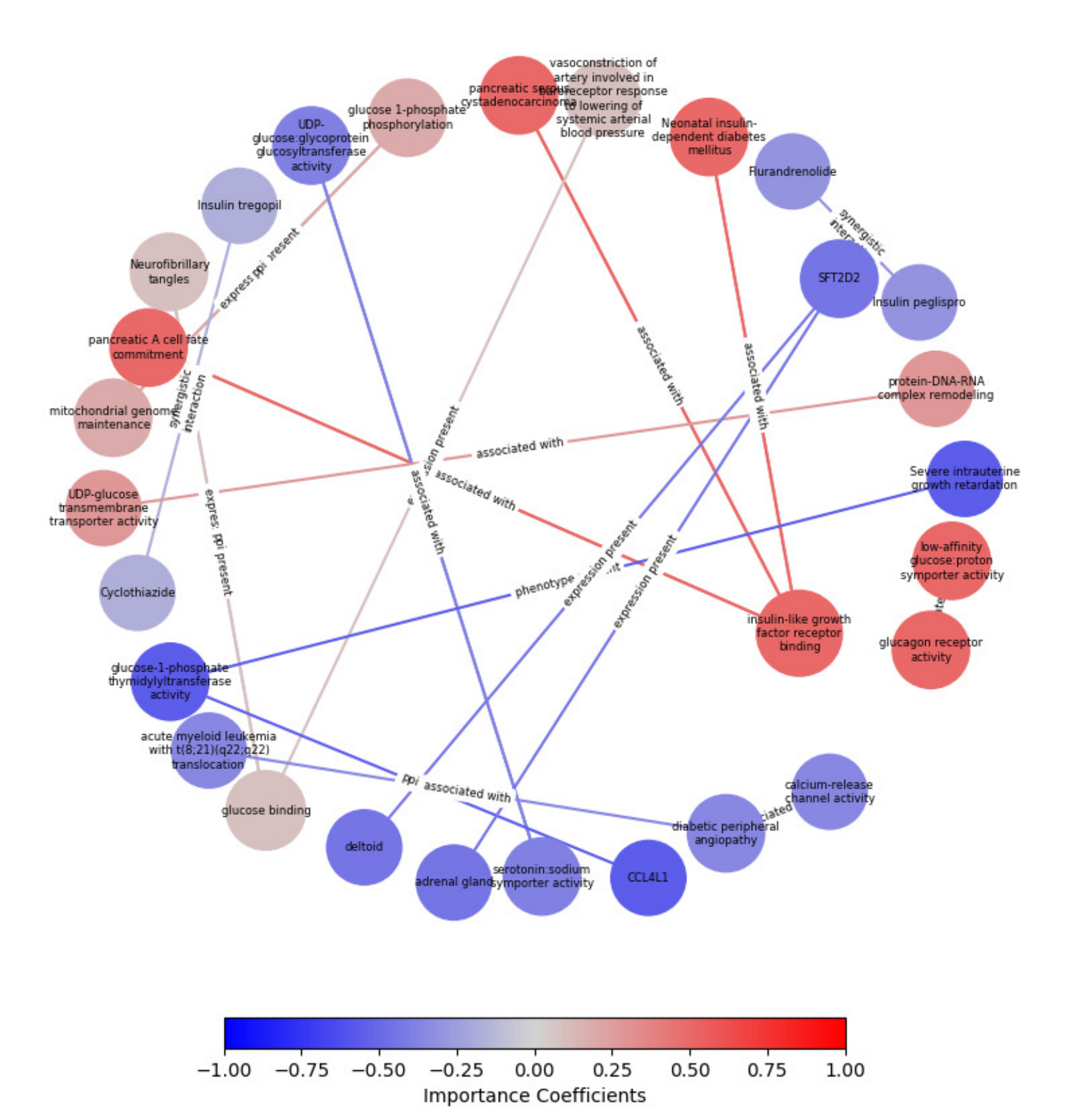}
    \caption{  Preprompt KG-SMILE explainability
    }
    \label{fig:Preprompt}
\end{figure*}
As illustrated in Fig.~\ref{fig:Preprompt},by integrating pre-prompting, selective response filtering, and answer aggregation, I significantly enhanced the stability, accuracy, and interoperability of KG-based QA systems. This approach ensures that even if an initial direct query results in \textit{"I do not know,"} alternative formulations allow the system to retrieve a meaningful response. While explicit node attribution may be more distributed, the modelstill correctly highlights the relevant knowledge components, making this technique particularly valuable for high-stakes applications such as healthcare research, institutional knowledge management, and decision-support systems~\cite{liang2021explainable}, where both precision and transparency are critical.

\subsection{Cross-Model Evaluation}
This section discusses cross-model evaluation to analyse the performance of responses generated by the GraphRAG system against two specialized medical LLMs named Mistral and MedAlpaca. The objective of performing cross-model evaluation is to investigate how well the responses generated by this GraphRAG align with responses generated by medical LLMs for the same 10 questions. This analysis provides insights into the strengths and limitations of our approach. This system uses a constrained dataset from PrimeKGQA. Mistral and MedAlpaca are often considered suitable corpora for assessing "ground truth" in biomedical querying. These two LLMs are trained and fine-tuned on domain-specific medical corpora. For each of the 10 questions, responses are generated independently using these models. To compare responses, similarities are computed using a composite metric containing semantic and concept similarity.  Mean Composite Similarity is an average score combining semantic (meaning-based), concept overlap (key ideas match), and content (word/structure match). The ± number represents the standard deviation, indicating how much scores vary, with higher values indicating greater inconsistency.
To assess semantic similarity, embedding similarity is calculated using Sentence Transformers. Concept overlap and content similarity measures are performed using key concept extraction and cosine similarity. Mean Semantic Similarity represents just the meaning-based part by ignoring exact words and focusing on ideas only.
Classifications were assigned based on similarity thresholds: 
very\_low ($<0.3$), 
low (0.3–0.5),
 medium (0.5–0.7), 
high ($>0.7$)
A small subset of the full PrimeKGQA is used to implement GraphRAG, which constrains retrieved information. This subset focuses on specific entity-relation pairs relevant to biomedical concepts, leading to potentially incomplete responses compared to the comprehensive knowledge encoded in Mistral and MedAlpaca. At the temperature of 0, GraphRAG operates deterministically, strictly fetching triplets from the KG. At temperature 1, the system allows partial integration of the LLM's internal knowledge, resulting in more elaborate responses. However, in our Graph-based RAG, outputs are highly dependent on the supplied KG edges and nodes. Whereas medical LLMs can generalize from trained patterns, they may not align precisely with PrimeKGQA's specific data. The medical LLMs possess broader clinical knowledge and alternative interpretations that differ from our subset of KG. Additionally, potential hallucinations in the medical LLMs could further contribute to mismatches, though their medical fine-tuning mitigates this to some extent.

\begin{table}[ht]
\centering
\renewcommand{\arraystretch}{1.2}
\begin{tabularx}{\textwidth}{>{\raggedright\arraybackslash}p{3cm} 
                                    >{\centering\arraybackslash}p{3cm} 
                                    >{\centering\arraybackslash}p{3cm} 
                                    >{\raggedright\arraybackslash}X}
\toprule
\textbf{Comparison} & \textbf{Mean Composite Similarity} & \textbf{Mean Semantic Similarity} & \textbf{Classification Distribution} \\
\midrule
\multicolumn{4}{l}{\textbf{Temperature 0}} \\
vs Mistral   & 0.319 ($\pm$0.214) & 0.515 & very\_low: 4 (40.0\%)\newline low: 3 (30.0\%)\newline medium: 3 (30.0\%) \\
vs MedAlpaca & 0.017 ($\pm$0.034) & 0.017 & very\_low: 10 (100.0\%) \\
\midrule
\multicolumn{4}{l}{\textbf{Temperature 1}} \\
vs Mistral   & 0.470 ($\pm$0.198) & 0.716 & very\_low: 3 (30.0\%)\newline low: 3 (30.0\%)\newline medium: 3 (30.0\%)\newline high: 1 (10.0\%) \\
vs MedAlpaca & 0.020 ($\pm$0.024) & 0.020 & very\_low: 10 (100.0\%) \\
\bottomrule
\end{tabularx}
\caption{Similarity statistics for GraphRAG outputs at different temperatures against medical LLMs.}
\label{tab:similarity-temp-clean}
\end{table}

The results in Table~\ref{tab:similarity-temp-clean} indicate low overall similarities to both Mistral and MedAlpaca. Considering KG-based constraints: at a temperature of 0, it is highly precise for KG triplets but lacks the depth of context-rich outputs of medical LLMs. For instance, in Question 1, GraphRAG at temp 0 Mistral provides a broader physiological explanation, whereas GraphRAG interprets relationships and entities strictly from the literal KG subset. Temperature 1 exhibits modest improvements (e.g., 70\% better than Mistral), as it leverages the LLM's internal knowledge to elaborate; however, it remains KG-dependent and thus diverges from the medical models' parametric generalizations. Temp 1 performs slightly better than Temp 0 against Mistral, as it utilizes AI knowledge, but has poor alignment with MedAlpaca. MedAlpaca results are poorer due to its clinically detailed focus, which prioritizes ethical, patient-oriented phrasing that is missing in our PrimeKGQA subset. The PrimeKGQA subset excludes comprehensive relations, resulting in "differing responses." 
The KG's sparsity restricts response depth, unlike the medical LLMs' broader knowledge. Despite this, the results are not entirely poor; the moderate similarity to Mistral (0.47 at Temp 1) and consistency (0.617 vs. MedAlpaca) suggest reliable outputs within the KG's scope, potentially comparable to human evaluation for fact-specific queries. The overall assessment results confirm that the PrimeKGQA subset is accurate and that the KG-based RAG is precise, but it needs a more comprehensive KG to broaden its coverage, marking a key direction for future work. Additional improvements could involve iterative KG expansion to capture emerging biomedical relations.

\subsection{Future Prospects}

GraphRAG technology faces several key challenges while also offering promising research opportunities. One area of focus is the development of dynamic and adaptive graphs instead of static databases ~\cite{edge2024local,fu2020survey,lan2021survey,lan2022complex,luo2023chatkbqa,yani2021challenges} that can incorporate real-time updates. Additionally, integrating multi-modal data such as images and audio would enhance knowledge representation~\cite{wei2019mmgcn}.

 Addressing scalability is crucial for managing large-scale KGs, which requires advanced retrieval mechanisms and efficient infrastructure. The use of graph foundation models [42, 115]can improve the processing of graph-structured data, while lossless compression techniques are necessary for effectively handling long contexts~\cite{lan2022complex,fu2020survey}

 Establishing standard benchmarks would create a framework for evaluating methodologies, promoting consistency and progress within the field. Expanding the applications of GraphRAG to complex domains, such as healthcare~\cite{kashyap2024knowledge}, finance~\cite{arslan2024business}, legal compliance~\cite{kim2025rag}, and the Internet of Things (IoT)~\cite{srivastava2020iot}, could significantly increase its impact.

 These challenges highlight the potential for GraphRAG to evolve into a robust and versatile technology across a variety of domains.

\section{Conclusion}\label{conclusion}

In conclusion, this study presents KG-SMILE as a framework that brings explainability and interpretability to GraphRAG systems. While standard RAG helps reduce hallucinations by grounding outputs in external knowledge, it still functions as a black box, users cannot see which retrieved pieces of information shaped the final response. By introducing perturbation-based attribution analysis, KG-SMILE makes this process more transparent: it systematically alters graph components and observes the effect on model outputs, thereby identifying the entities and relations most influential in generating a response. In this way, GraphRAG becomes not only a retrieval-and-generation pipeline but also an explainable system that can clarify \textit{why} an answer was produced.

This capacity for explanation is especially important in sensitive domains such as healthcare and law, where decision-making requires not only accurate outputs but also accountability and trust in the reasoning process. Rather than claiming higher performance, the emphasis here is on transparency and traceability, ensuring that users can understand the link between retrieved knowledge and generated answers. Looking forward, continued research into dynamic graph adaptation, multi-modal integration, scalability, and standardized evaluation benchmarks will further shape the role of explainable GraphRAG systems. With these directions, such frameworks can provide a foundation for responsible AI applications across various sectors, including healthcare, finance, IoT, and legal systems.

\appendix
\section{Additional Results}

To complement the results presented in the main text, we provide in 
Table~\ref{tab:Auc-table-biomedical} shows the area under the ROC curve (AUC) scores 
for biomedical question–answer pairs evaluated at two different decoding 
temperatures. These results illustrate the effect of temperature variation 
on answer consistency across a range of biomedical tasks.

\begin{table}[ht]
  \centering
  \renewcommand{\arraystretch}{1}
  \small
  \setlength{\tabcolsep}{6pt}
  \begin{tabularx}{\textwidth}{>{\raggedright\arraybackslash}X c c}
    \toprule
    \textbf{Question} & \textbf{AUC ($T{=}0$)} & \textbf{AUC ($T{=}1$)} \\
    \midrule
    What is insulin-like growth factor receptor binding associated with? & 1.00 & 0.78 \\
    What is pancreatic serous cystadenocarcinoma associated with? & 1.00 & 1.00 \\
    What is glucagon receptor activity associated with and how does it interact through ppi with low-affinity glucose:proton symporter activity? & 1.00 & 1.00 \\
    What is glucose binding? & 1.00 & 0.89 \\
    Which lifestyle changes are most effective in preventing neonatal insulin-dependent diabetes mellitus? & 0.50 & 0.44 \\
    What is glucose 1-phosphate phosphorylation and mitochondrial genome maintenance, and how do they interact through ppi? & 1.00 & 0.78 \\
    What is UDP-glucose:glycoprotein glucosyltransferase activity associated with, and how does it interact with serotonin:sodium symporter activity through ppi? & 0.89 & 0.78 \\
    Where is SFT2D2 expressed? & 0.89 & 0.78 \\
    What is diabetic peripheral angiopathy associated with? & 1.00 & 0.89 \\
    Which molecular interactions are associated with neonatal insulin-dependent diabetes mellitus? & 0.50 & 0.10 \\
    \bottomrule
  \end{tabularx}
  \caption{AUC (area under the ROC curve) at decoding temperatures $T{=}0$ and $T{=}1$; "ppi" denotes protein–protein interaction}
  \label{tab:Auc-table-biomedical}
\end{table}

\newpage
\section{Declarations}

\subsection{Funding}
No funding was received for conducting this study.

\subsection{Competing Interests}
The authors declare that they have no competing interests.

\subsection{Ethical Considerations}
his study did not involve human participants or animals; therefore, ethics approval and informed consent were not required.

\subsection{Data Availability}
The datasets generated and analysed during the current study are available in the PrimeKGQA repository, URL: \url{https://zenodo.org/records/13829395}, and DOI: \href{https://doi.org/10.5281/zenodo.13348626}{10.5281/zenodo . 13348626}.
https://github.com/koo-ec/XGRAG

\subsection{Author Contributions}
Zahra Zehtabi Sabeti Moghaddam: Conceptualization, methodology, analysis, and writing – original draft.\\
Zeinab Dehghani: Analysis, writing, and contributions to coding.\\
Maneeha Rani: Contributions to coding.\\
Dr. Koorosh Aslansefat: Supervision, methodology, review, and editing.\\
Dr. Bupesh Mishra: Review and editing.\\
Dr. Rameez Raja Kureshi: Review.\\
Prof. Dhaval Thakker: Supervision, Review.

\bibliography{sn-bibliography}  

\end{document}